\PassOptionsToPackage{unicode}{hyperref}
\PassOptionsToPackage{hyphens}{url}
\documentclass[
]{article}
\usepackage{lmodern}
\usepackage{amssymb,amsmath}
\usepackage{ifxetex,ifluatex}
\usepackage{authblk}
\ifnum 0\ifxetex 1\fi\ifluatex 1\fi=0 
  \usepackage[T1]{fontenc}
  \usepackage[utf8]{inputenc}
  \usepackage{textcomp} 
\else 
  \usepackage{unicode-math}
  \defaultfontfeatures{Scale=MatchLowercase}
  \defaultfontfeatures[\rmfamily]{Ligatures=TeX,Scale=1}
\fi
\IfFileExists{upquote.sty}{\usepackage{upquote}}{}
\IfFileExists{microtype.sty}{
  \usepackage[]{microtype}
  \UseMicrotypeSet[protrusion]{basicmath} 
}{}
\makeatletter
\@ifundefined{KOMAClassName}{
  \IfFileExists{parskip.sty}{%
    \usepackage{parskip}
  }{
    \setlength{\parindent}{0pt}
    \setlength{\parskip}{6pt plus 2pt minus 1pt}}
}{
  \KOMAoptions{parskip=half}}
\makeatother
\usepackage{xcolor}
\IfFileExists{xurl.sty}{\usepackage{xurl}}{} 
\usepackage[colorlinks=true, allcolors=blue]{hyperref}

\usepackage{caption}
\usepackage{graphicx,grffile}
\makeatletter
\def\maxwidth{\ifdim\Gin@nat@width>\linewidth\linewidth\else\Gin@nat@width\fi}
\def\maxheight{\ifdim\Gin@nat@height>\textheight\textheight\else\Gin@nat@height\fi}
\makeatother
\setkeys{Gin}{width=\maxwidth,height=\maxheight,keepaspectratio}
\makeatletter
\def\fps@figure{htbp}
\makeatother
\providecommand{\tightlist}{%
  \setlength{\itemsep}{0pt}\setlength{\parskip}{0pt}}
\usepackage[square,numbers]{natbib}
\title{\vspace{-3cm}Sensory Optimization: Neural Networks as a Model for Understanding and
Creating Art}

\author{Owain Evans\thanks{University of Oxford. \url{owaine@gmail.com}}}
\date{\normalsize November 2019}

\begin{document}
\maketitle
\begin{abstract}
\noindent This article is about the cognitive science of visual art. Artists
create physical artifacts (such as sculptures or paintings) which depict
people, objects, and events. These depictions are usually stylized
rather than photo-realistic. How is it that humans are able to
understand and create stylized representations? Does this ability depend
on general cognitive capacities or an evolutionary adaptation for art?
What role is played by learning and culture?

Machine Learning can shed light on these questions. It's possible to
train convolutional neural networks (CNNs) to recognize objects without
training them on any visual art. If such CNNs can \emph{generalize} to
visual art (by creating and understanding stylized representations),
then CNNs provide a model for how humans could understand art without
innate adaptations or cultural learning. I argue that Deep Dream and
Style Transfer show that CNNs \emph{can} create a basic form of visual
art, and that humans could create art by similar processes. This
suggests that artists make art by optimizing for effects on the human
object-recognition system. Physical artifacts are optimized to evoke
real-world objects for this system (e.g.~to evoke people or landscapes)
and to serve as superstimuli for this system.
\end{abstract}

{
\setcounter{tocdepth}{3}
\tableofcontents
}

\newpage 

\makeatletter
\newenvironment{chapquote}[2][2em]
  {\setlength{\@tempdima}{#1}%
   \def\chapquote@author{#2}%
   \parshape 1 \@tempdima \dimexpr\textwidth-2\@tempdima\relax%
   \itshape}
  {\par\normalfont\hfill--\ \chapquote@author\hspace*{\@tempdima}\par\bigskip}
\makeatother


\begin{chapquote}{E.H. Gombrich, \textit{Art and Illusion}}
The history of art \ldots{} may be described as the forging of master
keys for opening the mysterious locks of our senses to which only nature
herself originally held the key \ldots{} Like the burglar who tries to
break a safe, the artist has no direct access to the inner mechanism. He
can only feel his way with sensitive fingers, probing and adjusting his
hook or wire when something gives way.
\end{chapquote}
\label{epigraph}

\hypertarget{introduction}{%
\subsection{Introduction}\label{introduction}}

\begin{figure}
\centering
\includegraphics[width=0.9\textwidth,height=\textheight]{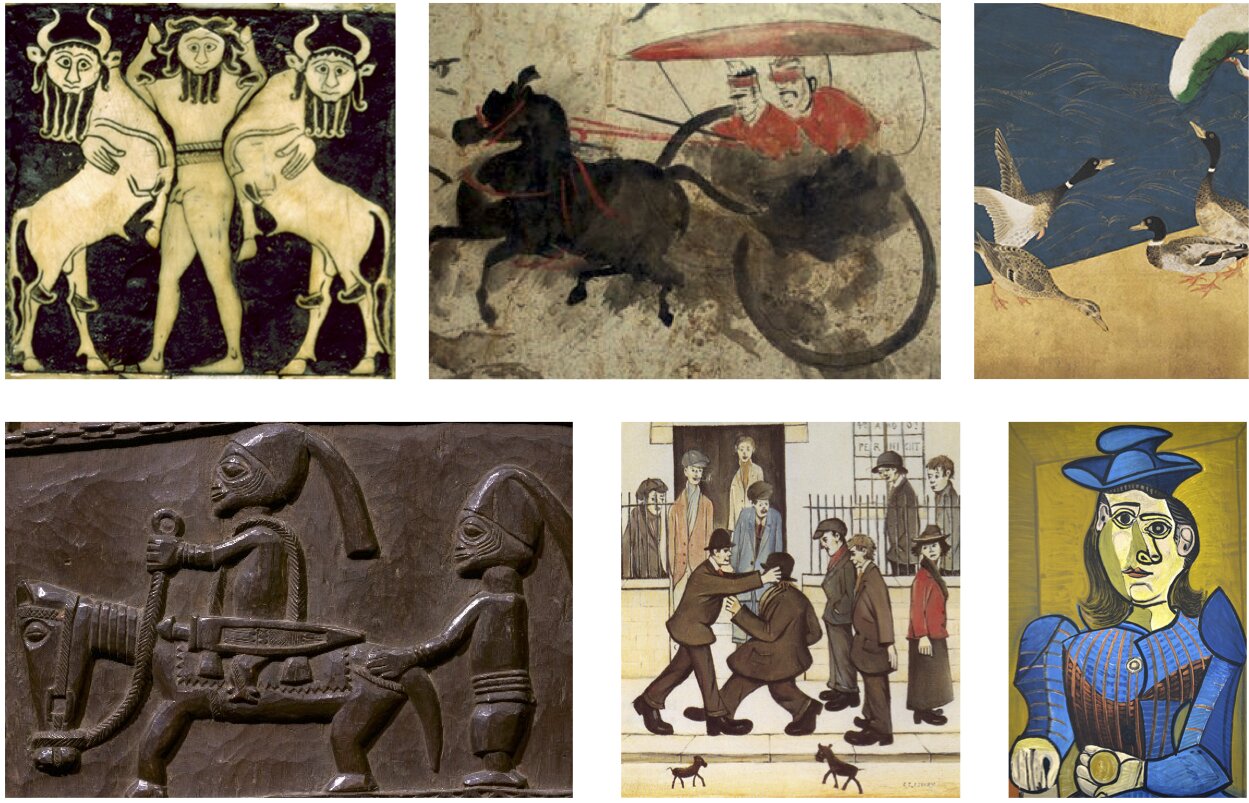}
\caption{Figure 1. Stylized depictions in visual art from different
cultures. Left-right: Sumer (2500BC), China (220AD), Japan (1700s),
Nigeria (1800s), UK (1935), France (1955).
\protect\hyperlink{sources}{Image sources.}}
\label{fig1}
\end{figure}

Look at the artworks in Figure \hyperref[fig1]{1}. Each of them is stylized and easily
distinguished from a photo. Yet you can effortlessly recognize what is
depicted (e.g.~people, horses, ducks) and form a judgment about the
aesthetic quality of the depictions.

This example raises some foundational questions about visual art and
human cognition:

\begin{itemize}
\item
  How do viewers recognize what is depicted in a stylized image and
  judge the aesthetic quality of the image?
\item
  How do artists create stylized representations that are easily
  recognizable even to viewers from a distant culture?
\item
  How did humans invent visual art in the first place?
\end{itemize}

Answers to these questions will invoke both general human abilities
(e.g.~vision, manual dexterity) and abilities specialized for visual
art. Yet what is the balance between them? Here are two opposing
positions:

\begin{quote}
\emph{Art-specific Position}\\
On this view, understanding stylized representations requires skills
specific to visual art. These skills have two very different sources.
Humans could have an evolutionary adaptation for art, an ``art
instinct'' analogous to the language instinct \citep{dutton2009art}.
Humans can also \emph{learn} art-specific skills. These skills could
come from immersion in visual art or from being taught an explicit
``visual language'', where arbitrary symbols are used to denote concepts
\citep{goodman1976languages, hyman2017depiction}.
\end{quote}

\begin{quote}
\emph{Generalist Position}\\
On this view, general human abilities are sufficient to understand and
create stylized representations. These abilities depend on both innate
and learned capacities (e.g.~learning to recognize animals) but have
nothing to do with art \citetext{\citealp[
pp.526]{pinker2003how}; \citealp{ramachandran1999science}}. This
predicts that humans with no exposure to art could understand Figure \hyperref[fig1]{1},
and that art is accessible across cultures because it exploits general
abilities shared by all humans. In Gombrich's \hyperref[epigraph]{analogy} artists
are like locksmiths: they forge ``master keys for opening the mysterious
locks of our senses'' \citep{gombrich1960art}.
\end{quote}

There have been some experimental tests of the Generalist position. In a
psychology study in the 1960s, two professors kept their son from seeing
any pictures or photos until the age of 19 months. On viewing
line-drawings for the first time, the child immediately recognized what
was depicted \citep{hochberg1962pictorial}.\footnote{This experiment was
  imperfect because the child had some brief, unintended exposure to
  visual representations.} Yet aside from this study, we have limited
data on humans with zero exposure to visual representations. The
Generalist position also suggests that there's no innate art instinct.
This is hard to determine today because neuroscience lacks a detailed
picture of how the visual system processes art.

Deep neural networks provide a new source of evidence for choosing
between the Art-specific and Generalist positions. For the first time in
history, there are algorithms for object recognition that approach human
performance across a wide range of datasets
\citep{zoph2018learning, litjens2017survey}. This enables novel
computational experiments akin to depriving a child of visual art. It's
possible to train a network to recognize objects (e.g.~people, horses,
chairs) without giving it any exposure to visual art and then test
whether it can understand and create artistic representations. In other
words, can a network trained to recognize ordinary objects generalize to
visual art? If so, this would support the Generalist position. This
``deprivation'' experiment for neural nets has not yet been carried out
systematically. However similar experiments have: people have used Deep
Dream and Style Transfer to generate intriguing images
\citep{mordvintsev2015inceptionism, gatys2016image}. In this article, I
will argue that the results of Deep Dream and Style Transfer show that
neural nets for object recognition \emph{can} generalize to art. This
helps to explain how human general abilities could suffice for
understanding and creating visual art, and so provides support for the
Generalist position.

Here is an outline of the article:

Part 1 reviews evidence that neural nets for object recognition can
generate art-like images without having been exposed to visual art
during training. I explain Feature Visualization, Deep Dream, and Style
Transfer. These techniques create images by optimizing to cause a
certain pattern of activation in a trained neural net. The techniques
can be combined to produce images that are \emph{superstimuli} for
features of the network (as in Feature Visualization), and
\emph{transcriptions} of content into a different style or medium (as in
Style Transfer).

In Part 2, I argue that humans could create art by a similar process to
Feature Visualization and Style Transfer. This helps to explain the
origins and development of visual art on the Generalist position. Humans
do not perform gradient descent but instead apply general intelligence
to optimize physical artifacts for the human visual system.

\hypertarget{part-1-creating-art-with-networks-for-object-recognition}{%
\subsection{Part 1: Creating art with networks for object
recognition}\label{part-1-creating-art-with-networks-for-object-recognition}}

Can neural networks trained to recognize objects generalize to visual
art (having been ``deprived'' of art during training)? Can such networks
\emph{create} visual art? In reviewing the evidence, I limit my
discussion to convolutional networks trained to label objects on
datasets like ImageNet \citep{deng2009imagenet}. Networks trained
explicitly to generate images, such as GANs and variational
autoencoders, are beyond the scope of this article.\footnote{GANs
  generate impressive artistic images but they seem less informative
  than ImageNet models about the Generalist vs.~Art-specific debate.
  First, some GANs are trained on human visual art, while ImageNet
  models are not. Second, GANs are explicitly trained to \emph{generate}
  images. We know the human visual system is optimized by evolution for
  object recognition but we are uncertain about whether it's also
  optimized for generating images \citep{chater2018mind}.}

The ImageNet dataset is not completely free of visual art. A small
proportion of images contain art or design.\footnote{For example, the
  ImageNet classes for jigsaw puzzles and books.} Yet it's not part of
the task to recognize what is depicted in this art, and so training on
ImageNet is a reasonable approximation of ``depriving'' a network of
art. A related concern is that the network learns something about visual
art simply by training on photos, because photos are created and
processed for human consumption. I discuss this concern in Section 2.4.

I will introduce some terminology to aid exposition. I refer to a
convolutional net trained on ImageNet as a ``recognition net''. The
experiments described in this article use a few different convolutional
architectures (VGG, GoogLeNet and ResNets) and so the term ``recognition
net'' will refer to one of these architectures.

\hypertarget{can-neural-nets-comprehend-human-paintings}{%
\subsubsection{1.1. Can neural nets comprehend human
paintings?}\label{can-neural-nets-comprehend-human-paintings}}

\begin{figure}
\centering
\includegraphics[width=1\textwidth,height=\textheight]{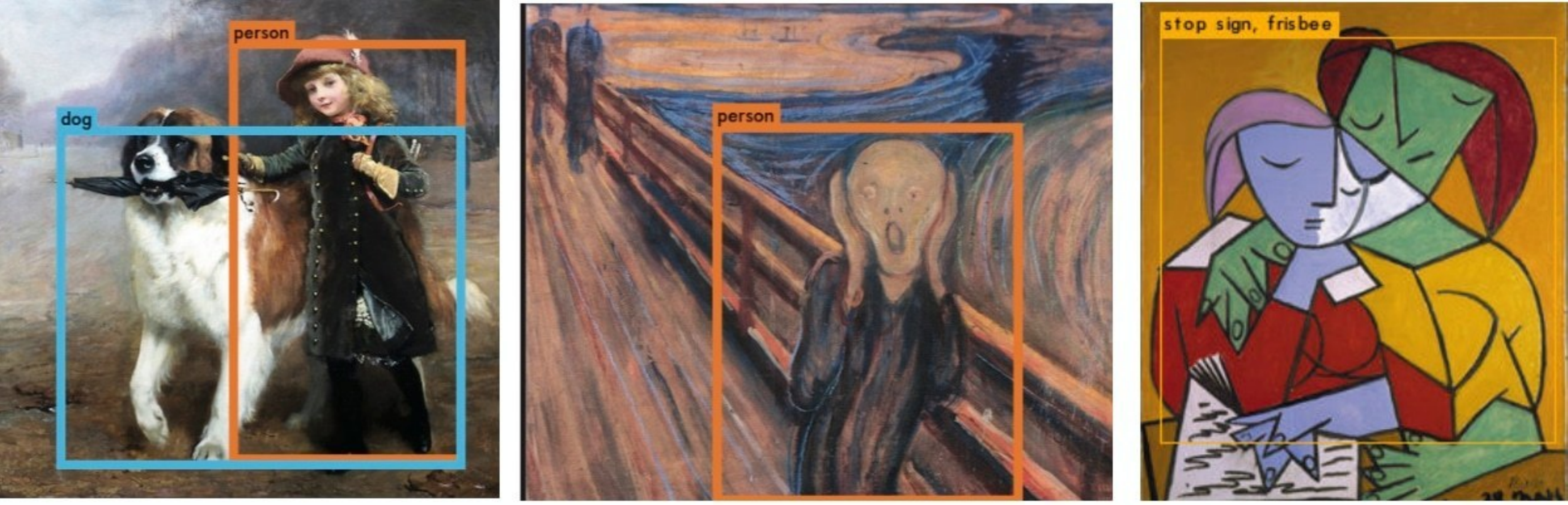}
\caption{Figure 2. Outputs from applying YOLO to paintings. YOLO is a
conv-net based model for recognition and localization. It was not
trained on visual art but can generalize to it. In the two leftmost
images, YOLO successfully recognizes humans and a dog (reproduced from
\citep{redmon2016you}). In the image on the right, YOLOv3 incorrectly
identifies people as ``stop sign'' or ``frisbee'' (generated by the
author using \citep{redmon2018yolov3}).}
\label{fig2}
\end{figure}

The simplest test for whether a recognition net can comprehend visual
art is to run the net on paintings and drawings. If there's a dog in the
painting, is it recognized as a dog? Unfortunately, there is not much
research on this question. One paper tested YOLO, a conv-net based
model, on a range of European paintings \citep{redmon2016you}. Figure \hyperref[fig2]{2}
shows some of the network's outputs. The overall results are fairly
impressive but substantially below human performance. The results could
likely be improved by applying recent advances in visual recognition.

\hypertarget{generating-images-by-feature-visualization}{%
\subsubsection{1.2. Generating images by Feature
Visualization}\label{generating-images-by-feature-visualization}}

\begin{figure}
\centering
\includegraphics[width=0.9\textwidth,height=\textheight]{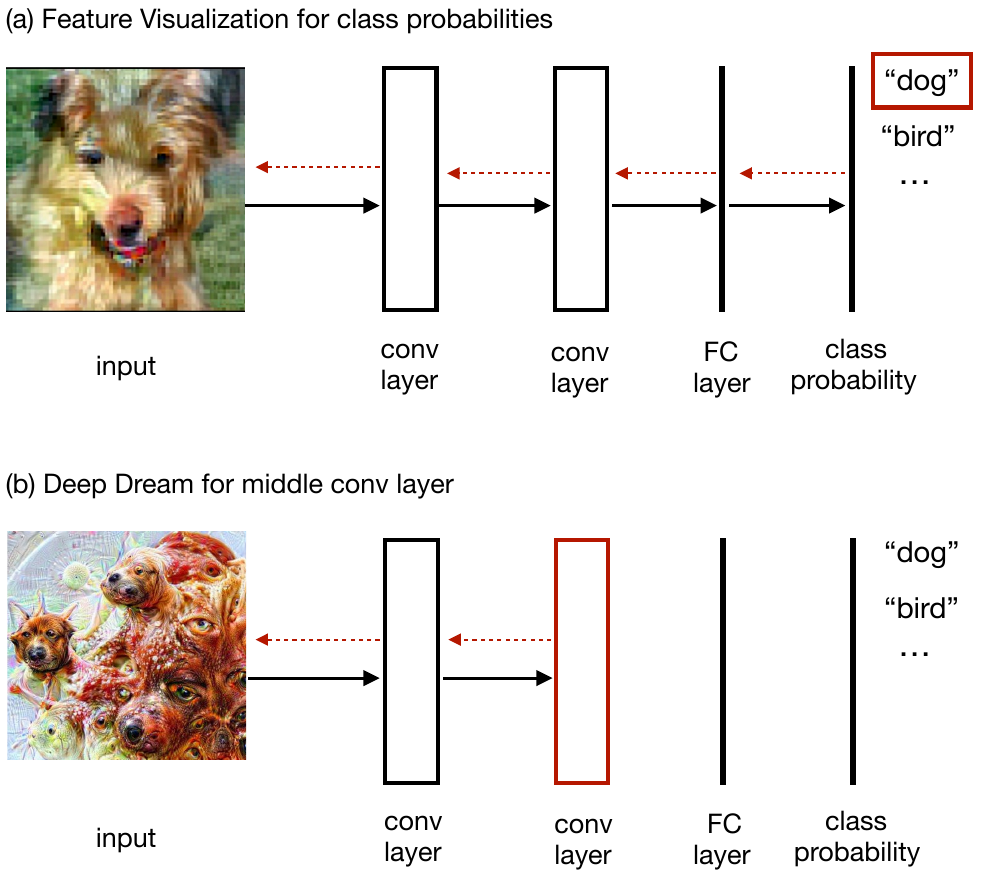}
\caption{Figure 3. Schematic diagrams of Feature Visualization and Deep
Dream for convolutional nets. Diagram (a) shows FV for class
probabilities/labels. At each timestep, the class probability is
computed for the image (black arrows) and the image is updated using the
gradient of the probability (red arrows). Diagram (b) shows DD, where
the objective is the squared activation of one of the conv layers.}
\label{fig3}
\end{figure}

One way of using a recognition net to generate art-like images is
Feature Visualization (``FV'') and the closely related Deep Dream.
Researchers developed FV to help interpret neural networks, by
visualizing the features computed by neurons
\citep{olah2017feature, nguyen2016multifaceted}. To visualize a neuron
we synthesize an image that maximizes the neuron's activation (Fig.~\hyperref[fig3]{3}).
The image pixels are iteratively optimized by gradient ascent,
backpropagating the activation from the neuron to the pixels. The same
technique can be used to visualize a class label (e.g.~the ``fox''
class), a channel, or an entire layer (as in Deep Dream). Intuitively,
the idea of FV is to create an image that's a \emph{superstimulus} for a
neuron without knowing in advance what feature the neuron corresponds
to.

I will sketch the formal details of FV. Suppose we are generating a
visualization for the \(i\)-th class label. Let \(f_\theta^i(x)\) be the
probability that image \(x\) is in class \(i\), where \(f_\theta\) is
the neural net with parameter vector \(\theta\) and \(x\) is a vector of
the image pixels. The objective is to find the image \(x^*\) such that:

\[ x^* = \underset{x}{\mathrm{argmax}}{ ( f_\theta^i(x) )}\]

To optimize this objective, we compute the derivative of the probability
with respect to the components of \(x\):

\[\frac{ \partial(f_\theta^i(x)) }{ \partial x }\]

This contrasts with the normal use of backpropagation for training the
network, where the derivative is taken with respect to components of
\(\theta\). To visualize a neuron in one of the middle convolutional
layers, we evaluate \(f_\theta\) up to that layer and backpropagate from
there. For initialization, the image \(x\) is set either to random noise
or to a photo. Projected Gradient Ascent can be used to keep \(x\) from
diverging too much from the photo \citep{santurkar2019computer}, and
image regularization and preconditioning can be used to reduce
high-frequency noise \citep{olah2017feature}.

\begin{figure}
\centering
\includegraphics[width=0.6\textwidth,height=\textheight]{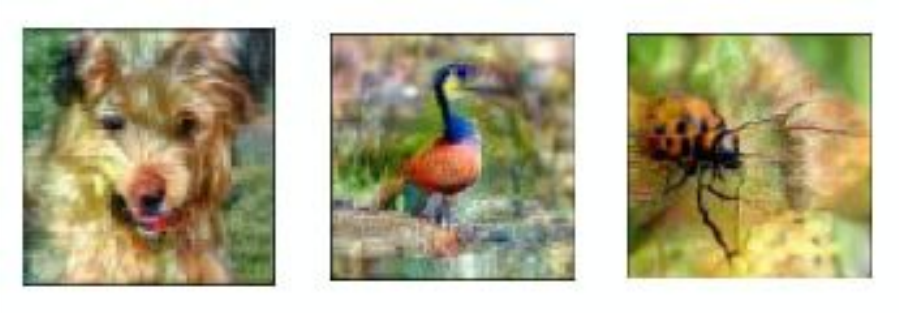}
\caption{Figure 4. Visualizations for class labels ``dog'', ``bird'',
and ``insect'' for a robust ResNet trained on Restricted ImageNet.
Images are initialized to a Gaussian fit to training photos and
optimized by PGD. Reproduced from \citep{santurkar2019computer}}
\label{fig4}
\end{figure}

Visualizations for the class labels of a robustly-trained ResNet are
shown in Figure \hyperref[fig4]{4} \citep{santurkar2019computer}. It's an important and
surprising fact that these images resemble animals. The recognition net
was trained only to \emph{distinguish} between animals, not to generate
images of them.\footnote{In principle it's possible to distinguish
  objects without being able to generate pictures of them: e.g.~zebras
  can be distinguished from horses by looking for stripes.} Yet the net
implicitly learns a model of how the animals look, which FV is able to
extract and visualize.

\begin{figure}
\centering
\includegraphics[width=0.85\textwidth]{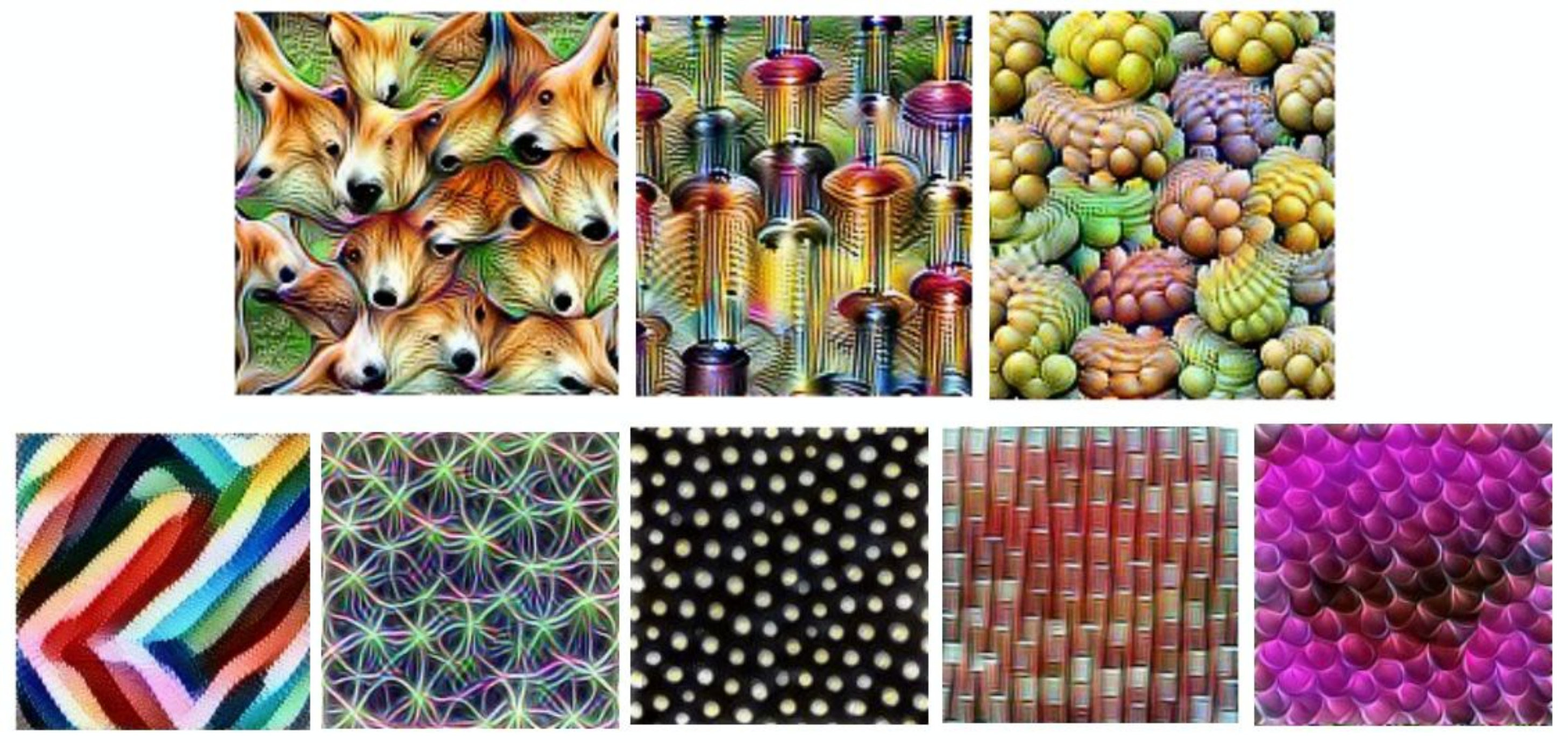}
\caption{Figure 5. Top row: Feature Visualizations for channels in a
middle convolutional layer of GoogleNet, which seem to code for pointy
dog snouts, the top of screws, and bunches of fruit. Bottom row:
Visualizations for channels in an earlier convolutional layer, coding
for curved diagonal lines, webbing, spots, grid squares, and pink
fabric. Images reproduced from \citep{olah2017feature}.}
\label{fig5}
\end{figure}

Visualizations for neurons from the middle layers of a recognition net
(GoogleNet) are shown in Figure \hyperref[fig5]{5} \citep{olah2017feature}. These neurons
seem to code for parts of objects that are useful for classification,
such as pointy snouts and the top of a screw. The visualizations look
quite different from photos in ImageNet. Like the superstimuli studied
in animal behavior research, they are simplified, off-distribution
inputs that cause intense activation for the network
\citep{superstimulus, ramachandran1999science}.

Neurons in earlier layers seem to compute lower-level features of
objects (Fig.~\hyperref[fig5]{5}). Visualizations for these neurons contain abstract,
geometric forms, including:

\begin{itemize}
\tightlist
\item
  blocks of color
\item
  straight lines, V-shapes, zigzags, circles.
\item
  grids, honeycomb tiling, spots, webs
\end{itemize}

These forms are also found in abstract and decorative art. Indeed,
various scientists have hypothesized that people enjoy abstract forms in
art because they are superstimuli for geometric features the brain uses
to recognize objects.\footnote{This hypothesis is discussed by
  \citep{pinker2003how, dennett2006breaking, ramachandran1999science}.
  It's worth noting that neuroscience does not have a full picture of
  how low-level features are activated by visual art or why such
  activation would cause pleasure \citep{davies2012artful}.}

\hypertarget{deep-dream-caricatures-and-hybrids}{%
\subsubsection{1.2.1. Deep Dream, caricatures, and
hybrids}\label{deep-dream-caricatures-and-hybrids}}

Deep Dream (``DD'') is a variation on Feature Visualization where an
image is optimized for the activation of an entire convolutional layer
\citep{mordvintsev2015inceptionism}.\footnote{More precisely, the
  objective is to maximize the squared activation of all neurons in a
  layer.} Images generated by Deep Dream reflect the distinctions the
network cares most about. Training on ImageNet, which requires
distinguishing between 120 different dog breeds, produces a
preponderance of dogs, whereas the ``Places'' dataset produces arches
and windows.

\begin{figure}
\centering
\includegraphics[width=0.65\textwidth,height=\textheight]{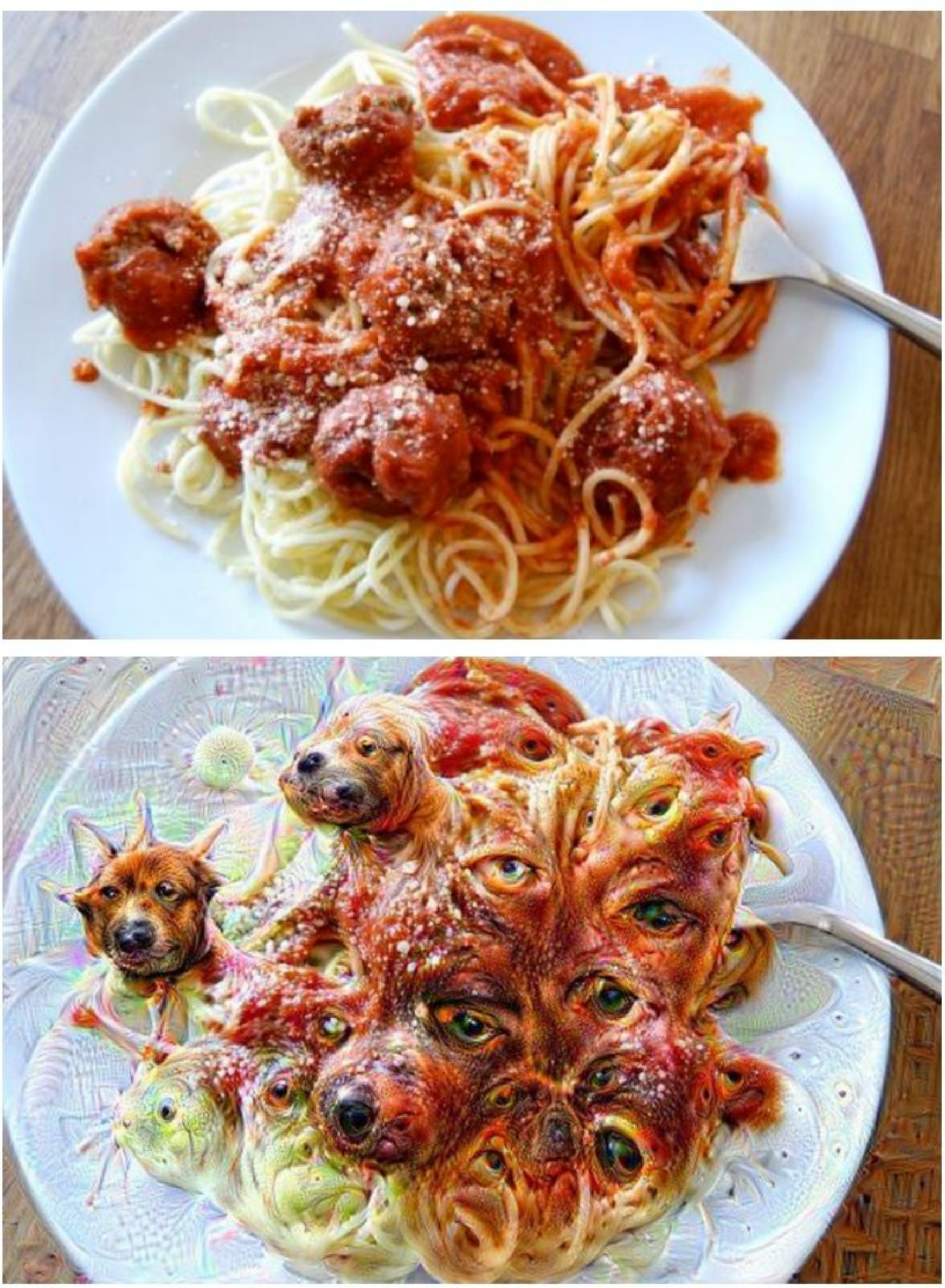}
\caption{Figure 6. Deep Dream applied to meatballs.}
\label{fig6}
\end{figure}

\begin{figure}
\centering
\includegraphics[width=0.8\textwidth,height=\textheight]{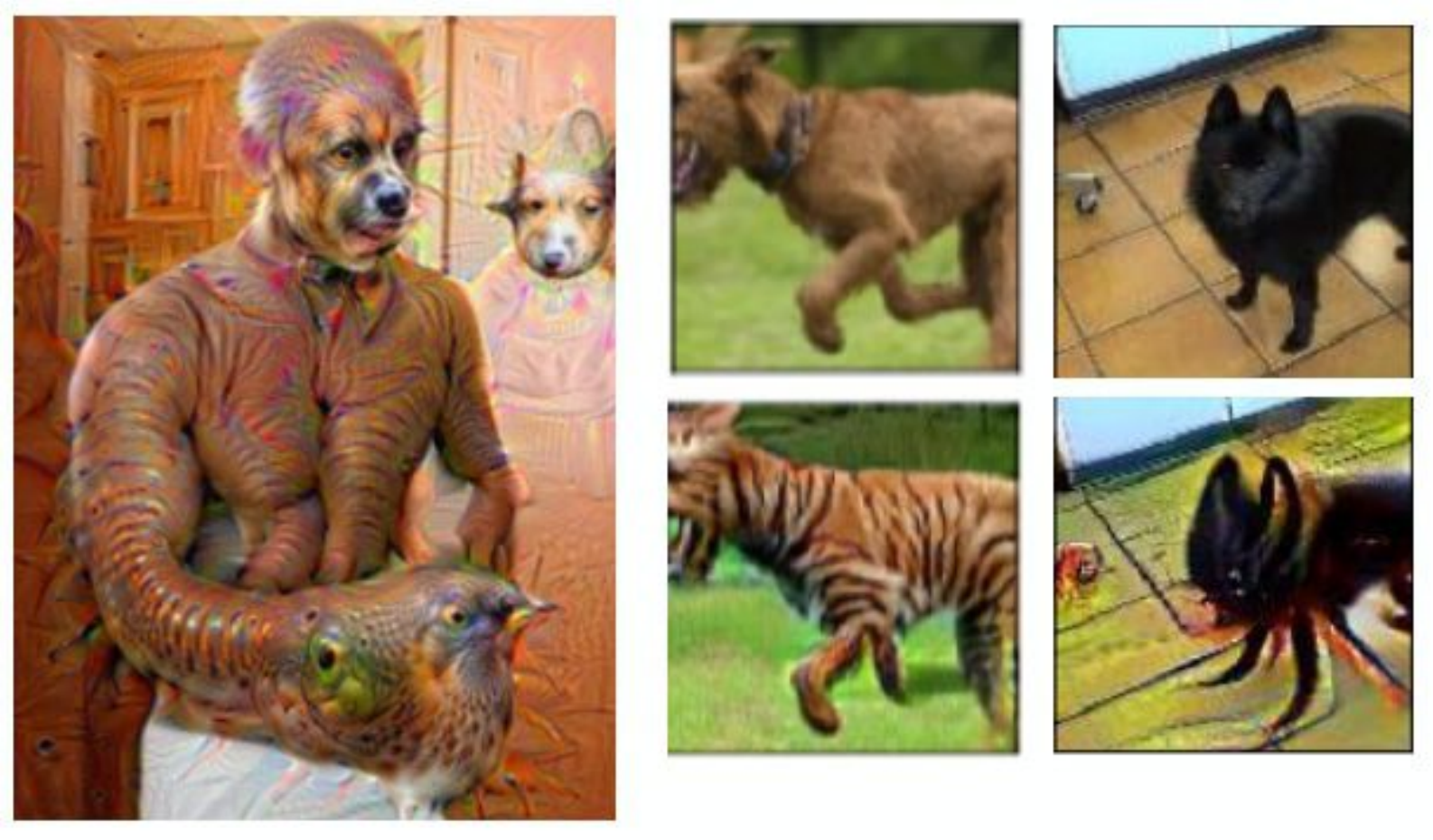}
\caption{Figure 7. Left: Deep Dream applied to a photo of a man. Right:
Using FV to add stripes or insect legs to animals. The image is
optimized for a particular neuron in the FC layer of a robust ResNet.
Images reproduced from \citep{engstrom2019learning}.}
\label{fig7}
\end{figure}

Figure \hyperref[fig6]{6} shows how Deep Dream transforms meatballs into dog heads. This
transformation is fairly subtle: if you zoom out the two images look
very similar. Yet there's a big difference in our experience of the
images. The transformed image is dense with details like eyes, scales,
and insect legs. Deep Dream can do a lot with a little, using subtle
changes to evoke an abundance of semantic detail. In this way, Deep
Dream is similar to stylized representations in visual art, such as
caricatures, which use a small amount of visual information to evoke a
particular individual \citep{olah2018github}. The artist John Ambrosi
has made spectacular use of this ``subtle'' application of Deep Dream in
his series ``Dreamscapes''.\footnote{See this cathedral
  \href{https://www.danielambrosi.com/Dreamscapes-2/i-ZqMVb3M/A}{image}
  and this nebula
  \href{https://www.danielambrosi.com/Dreamscapes-2/i-kSK9kHH/A}{image},
  as well as Ambrosi's entire
  \href{https://www.danielambrosi.com/Dreamscapes-Portal}{collection}.}
On the other hand, if Deep Dream is run for longer on an image, the
transformation becomes much less subtle (see Fig.~\hyperref[fig7]{7} left).

Deep Dream generates bizarre structures that look like hybrid creatures.
This kind of hybrid can also be created in a controlled way by FV, by
optimizing a photo towards particular features in the recognition net
(Fig.~\hyperref[fig7]{7}). It seems that for neural nets, the ability to generate normal
animals goes hand-in-hand with the ability to recombine animal parts
into hybrids. The same is plausibly true of humans, as hybrid creatures
are a common trope in visual art.

\hypertarget{deep-dream-art-for-the-neural-net}{%
\subsubsection{1.2.2. Deep Dream = art for the neural
net}\label{deep-dream-art-for-the-neural-net}}

The images generated by Feature Visualization and Deep Dream have some
of the basic properties of human visual art, including accurate
representation of natural objects, abstract forms, and caricature. Yet,
as I will discuss in Part 2, the images lack global coherence, emotional
expressiveness, and variety. My goal in this article is to relate neural
networks to questions of how humans understand and create visual art.
For this goal, the \emph{process} of Feature Visualization is as
important as the results. The process generates intense or
``meaning-dense'' stimuli for the recognition net. We could think of
this as art \emph{for the recognition net}, rather than art for humans.
It's noteworthy that humans also find Deep Dream images intense and
semantically dense. If the recognition net was trained on a task closer
to human learning (e.g.~captioning photos of complex social situations),
this ``art for neural nets'' would plausibly get closer to human visual
art (see Section 2.3).

\hypertarget{style-transfer-and-medium-transfer}{%
\subsubsection{1.3. Style Transfer and Medium
Transfer}\label{style-transfer-and-medium-transfer}}

There are two other properties of human visual art that Feature
Visualization does not capture:

\begin{enumerate}
\def\labelenumi{\arabic{enumi}.}
\item
  \emph{Style Re-use}\\
  Artists create new works that re-use the style of previous works.
\item
  \emph{Transcription}\\
  Humans make art in physical media with different formal properties
  than human visual perception. For example, woodcut prints are 2D,
  monochrome, and static, whereas human visual perception is a
  binocular, full-color video stream. (By contrast, FV generates images
  with the same form as ImageNet photos). So artists \emph{transcribe}
  their visual perception into physical media with formal constraints.
\end{enumerate}

\begin{figure}
\centering
\includegraphics[width=0.9\textwidth,height=\textheight]{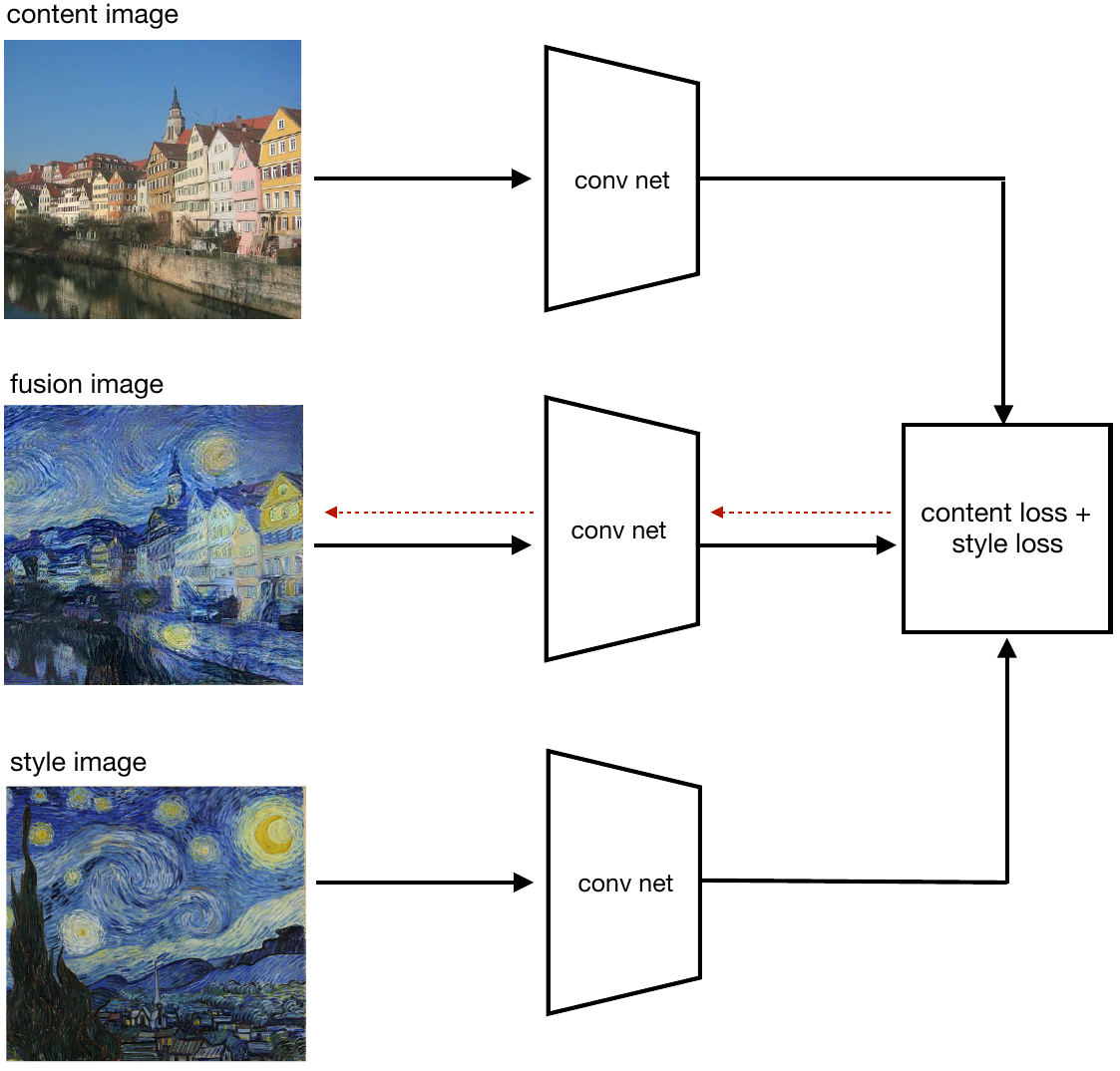}
\caption{Figure 8. Style Transfer. The \emph{representation} of each
image under the recognition net (``conv net'') is computed. The
representations are compared using the content and style loss functions
and backpropagated to the fusion image. Diagram adapted from
\citep{mordvintsev2018differentiable, gatys2016image}.}
\label{fig8}
\end{figure}

Neural Style Transfer (``ST'') is a technique that uses a recognition
net to achieve a simplified version of properties 1 and 2. ST generates
an image that fuses the style and content of two source images
\citep{gatys2016image}. Like FV, the algorithm for ST optimizes an image
to cause a particular response in a recognition net. Yet instead of
optimizing for the activation of a neuron, ST optimizes the image to
match the internal representation (or ``embedding'') of the style and
content images under the recognition net (see Figure \hyperref[fig8]{8}).

I will sketch the algorithm. As before, the recognition net \(f_\theta\)
is a convolutional net (such as VGG or robust ResNet) trained on
ImageNet. The \emph{representation} of an image \(x\) under \(f_\theta\)
at the \(k\)-th layer is defined as the set of all activations at the
layer. The \emph{style} of image \(x\) under \(f_\theta\) is defined as
a set of spatial statistics (the correlation matrix) of the
representations of \(x\) at multiple convolutional layers. Thus style
depends on low- and mid-level features of \(x\) but is invariant to
spatial location. The \emph{content} of \(x\) is its representation at
one of the later convolutional layers.\footnote{For full details,
  including the choice of which convolutional layers to get
  representations from, see \citep{gatys2016image}} Style Transfer aims
to generate a \emph{fusion image} \(x^*\) with style close to the style
image \(x_s\) and content close to the content image \(x_c\). Thus the
objective is a weighted sum of the \emph{style loss} \(L_c\) (the
\(L_2\) distance between styles) and the content loss \(L_c\) (the
\(L_2\) distance between contents):

\[ x^* = \underset{x}{\mathrm{argmin}}{( \alpha L_c(x,x_c) + \beta L_s(x,x_s))}\]

\begin{figure}
\centering
\includegraphics[width=1\textwidth,height=\textheight]{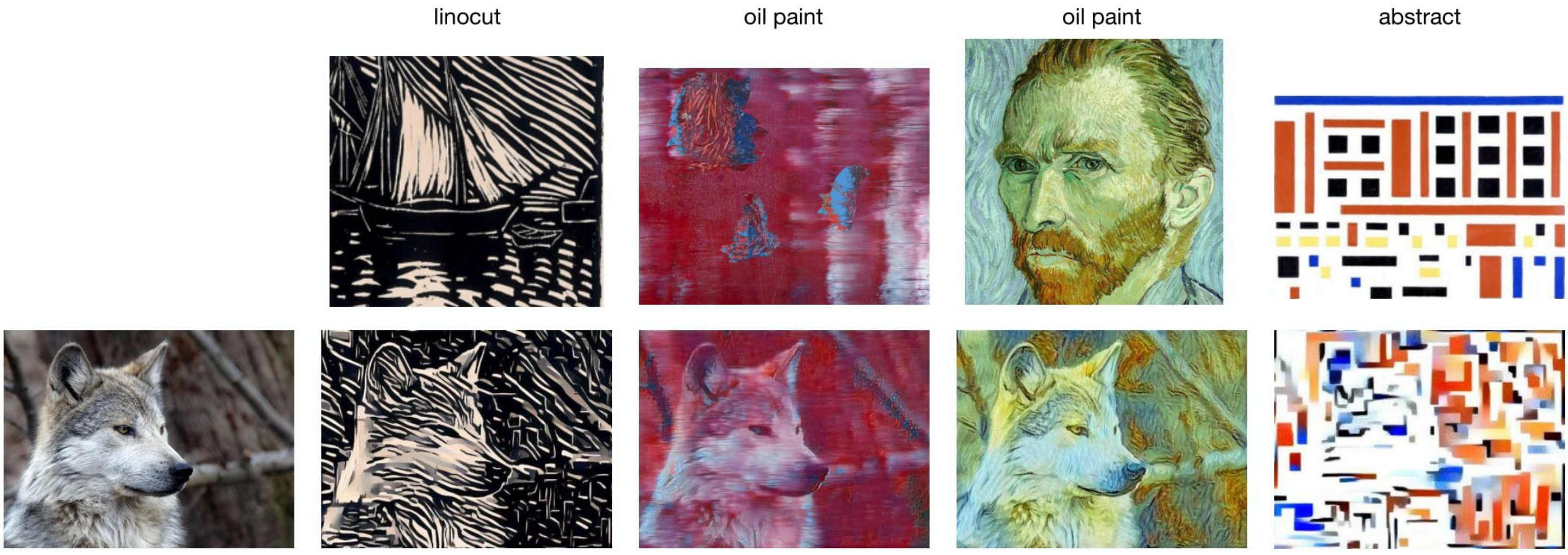}
\includegraphics[width=1\textwidth,height=\textheight]{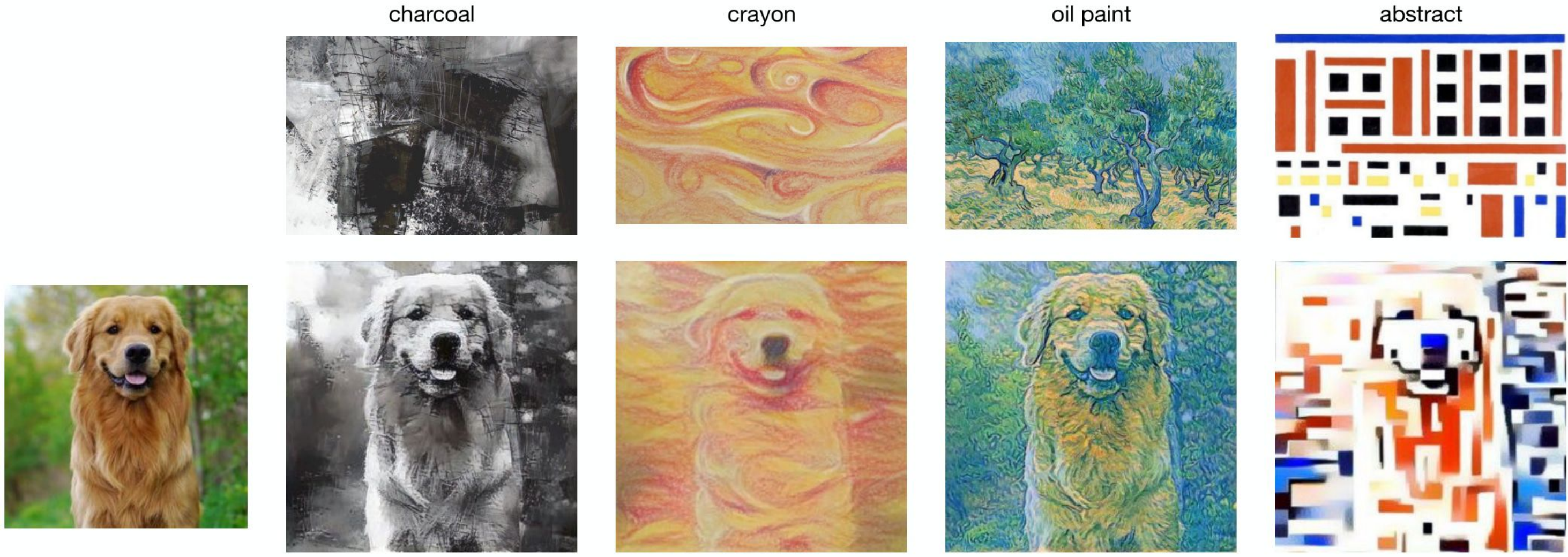}
\caption{Figure 9. Style Transfer. The style images have contrasting
low-level features and correspond to different physical media. This can
also be seen as a simplified version of ``Medium Transfer''. The fusion
images were generated by the author using VGG as in
\citep{gatys2016image}, initialized from the content image.
\protect\hyperlink{sources}{Image sources.}}
\label{fig9}

\centering
\includegraphics[width=0.8\textwidth,height=\textheight]{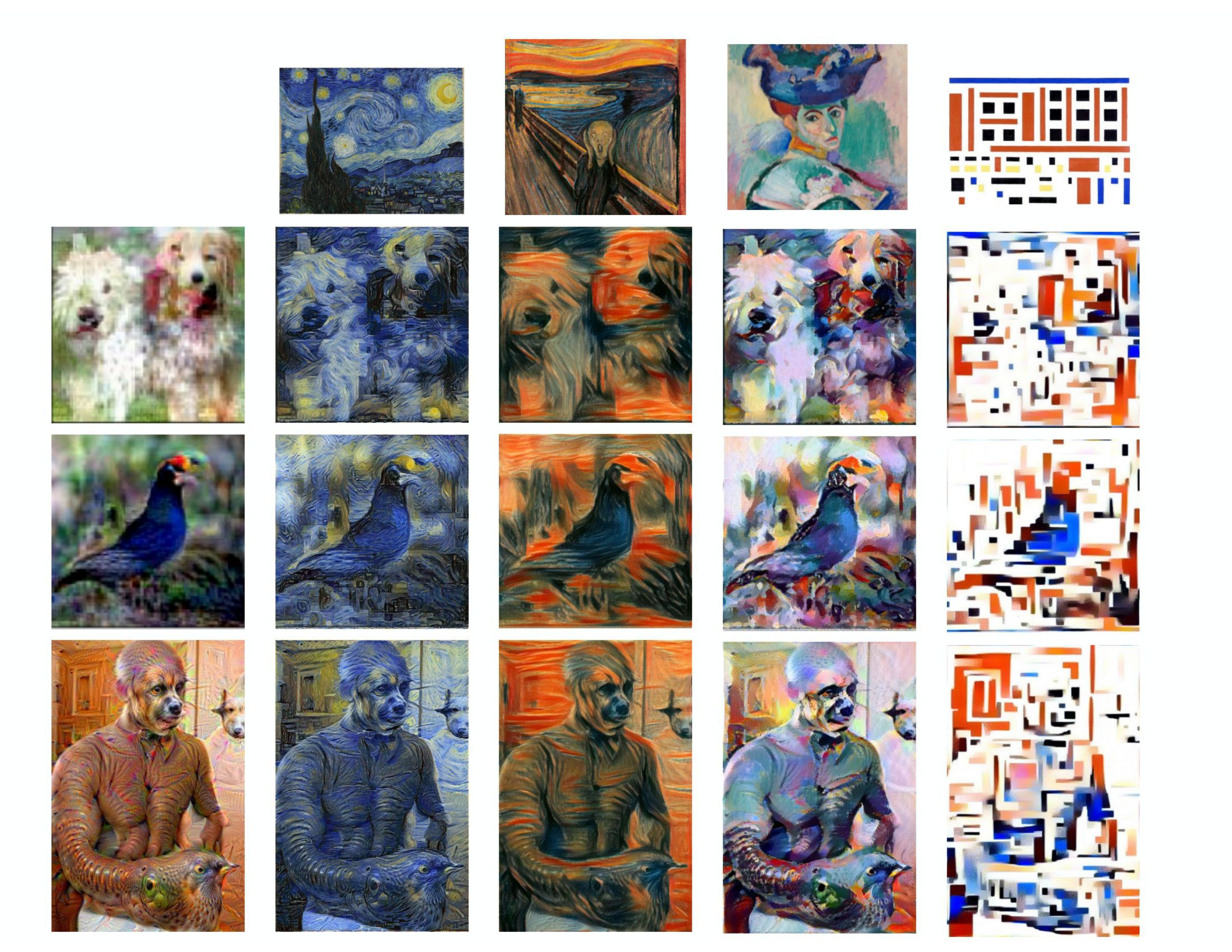}
\caption{Figure 10. Combining ST and FV.
The images in left column are content images generated by FV (first
two rows) and DD (third row). Fusion images generated by the author
using ST as in Fig.~9. Zoom in to see details.}
\label{fig10}
\end{figure}

Here \(\alpha\) and \(\beta\) are the weights for the style and content
losses. The definition of style in ST mostly captures small-scale
features like colors, textures, and brush strokes, and does not fully
capture the richer notion of style found in the study of art history.
Nevertheless, the results of ST are surprisingly impressive (Fig.~\hyperref[fig9]{9}) and
demonstrate two facts about the recognition net:

\begin{enumerate}
\def\labelenumi{\arabic{enumi}.}
\item
  By training on photos, the net has learned features that are
  general-purpose enough to capture the low-level structure of
  paintings.
\item
  The net can recognize the content even if the low-level textures
  differ from anything seen in training. For ST to generate Figure \hyperref[fig9]{9},
  the net must see the content (i.e.~a frontal shot of a dog) in the
  fusion images, despite never having seen a dog with the textures and
  colors of the ``crayon'' or ``abstract'' style images.\footnote{This
    might seem trivial, as the fusion image is optimized precisely to be
    recognizable to the net. However, the fusion images do not appear to
    be ``adversarial'' \citep{goodfellow2014explaining} as we humans can
    also recognize them as satisfying the objective of fusing the style
    and content images.}
\end{enumerate}

\begin{figure}
\centering
\includegraphics[width=.95\textwidth,height=\textheight]{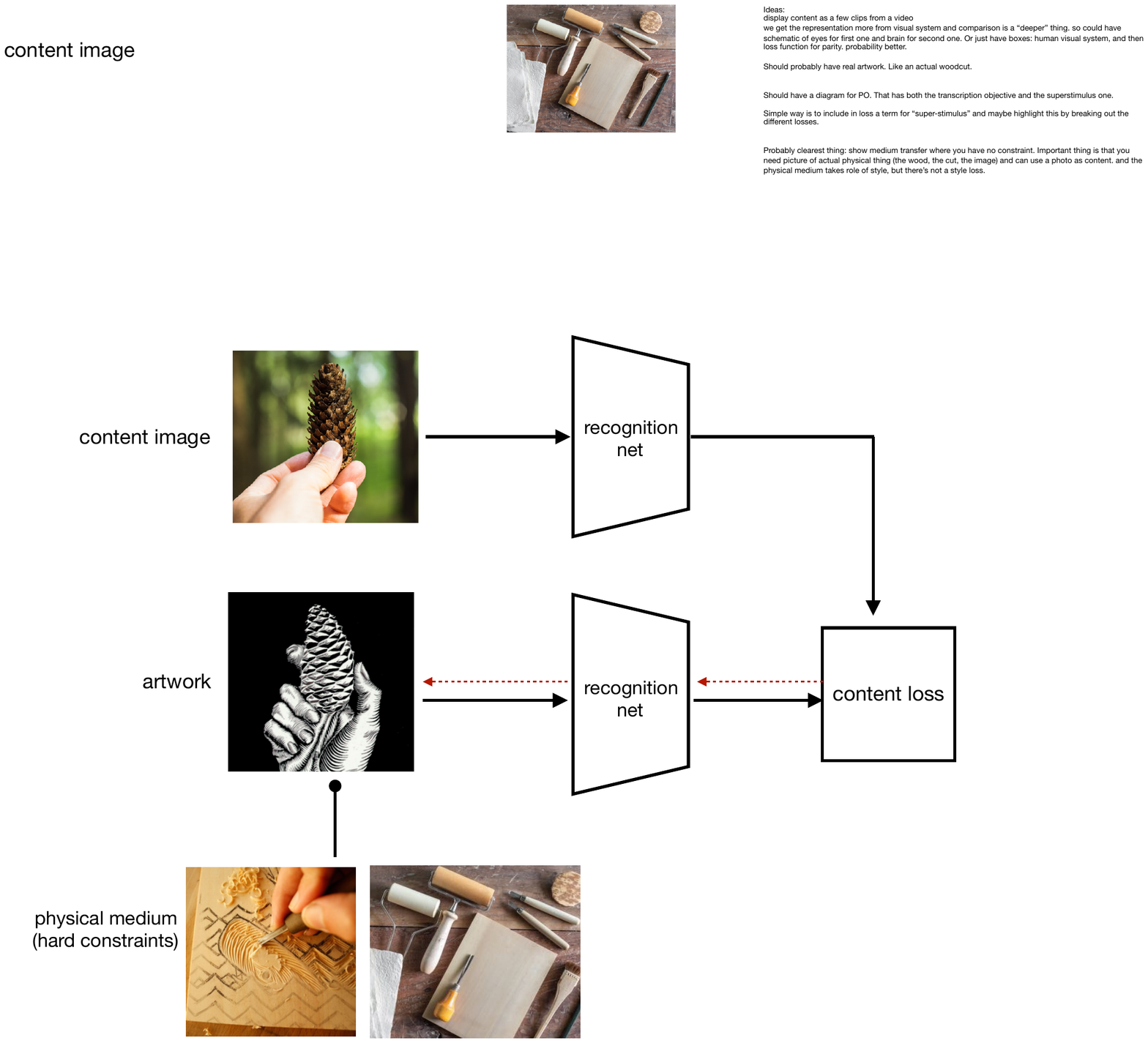}
\caption{Figure 11. Medium Transfer. A possible extension of ST where
the style loss is replaced by the hard constraints imposed by a physical
medium. In this example, the objective is to make the woodcut print that
best matches the content image. The print is created by making cuts into
wood and so is constrained to be monochrome and have low spatial
resolution. \protect\hyperlink{sources}{Image source.}}
\label{fig11}
\end{figure}

\hypertarget{the-physical-medium-of-visual-art}{%
\paragraph{1.3.1. The physical medium of visual
art}\label{the-physical-medium-of-visual-art}}

The style images for Figure \hyperref[fig9]{9} are taken from artworks in different
physical media, such as linocut, charcoal drawing and oil painting.
Style Transfer is able to transcribe content into fusion images that
reflect some of the low-level features of these media. This could be
taken a step further by designing physical artifacts rather than digital
images. The idea is to replace the soft constraints imposed by the style
image by hard constraints imposed by the physical medium (Fig.~\hyperref[fig11]{11}). The
physical artifact would be optimized such that, when viewed by the
recognition net, it causes a content representation similar to the
source content. I refer to this extension of ST as ``Medium Transfer''.
To implement Medium Transfer, you need to define the space of ways to
modify the physical medium. This would be the space of cuts for a
woodcut print (Fig.~\hyperref[fig11]{11}) or the space of arrangements of tiles in a
mosaic. Some physical artifacts have been designed using ideas similar
to Medium Transfer: see Section 1.4 (Perception Engines) and
\citep{mordvintsev2018differentiable, brown2017adversarial}.

Before computers made digital art possible, human artists had no choice
but to work in a physical medium. Yet physical media are also important
because they contribute to the viewer's experience of an
artwork.\footnote{The physical medium and the ``message'' or
  significance of the artwork can interact. Here are some examples: (1)
  The use of real gold in religious art. (2) A marble sculpture
  (e.g.~Bernini's
  \href{https://en.wikipedia.org/wiki/The_Rape_of_Proserpina}{Rape of
  Proserpina}) can be especially effective by accurately rendering skin
  or hair using a hard material. (3) In Degas's
  \href{https://commons.wikimedia.org/wiki/File:After_the_Bath,_Woman_drying_herself_-_Edgar_Degas_-_National_Gallery.jpg}{drawings}
  of a woman drying herself with towel, the ``trace of the pastel
  rubbing against the paper's surface subliminally registers the motion
  of the towel against the woman's skin'' \citep{hyman2017depiction}.}


\hypertarget{sensory-optimization-feature-visualization-style-transfer}{%
\subsubsection{1.4. Sensory Optimization = Feature Visualization + Style
Transfer}\label{sensory-optimization-feature-visualization-style-transfer}}

Feature Visualization and Style Transfer both optimize an image to cause
a particular pattern of activation in a recognition net. The two
objective functions can be combined into a single objective and
optimized in a unified process. I will introduce the term ``Sensory
Optimization'' (SO) for this kind of process. In particular, an SO
process has the following properties:

\begin{enumerate}
\def\labelenumi{\arabic{enumi}.}
\item
  Images are created by local optimization to cause a certain pattern of
  activation in a visual recognition system.
\item
  \emph{Superstimulus property}: images cause higher activation for
  features of the recognition system than training images.
\item
  \emph{Transcription property}: images are in a different physical
  medium and style than the training images.
\end{enumerate}

\begin{figure}
\centering
\includegraphics[width=0.9\textwidth,height=\textheight]{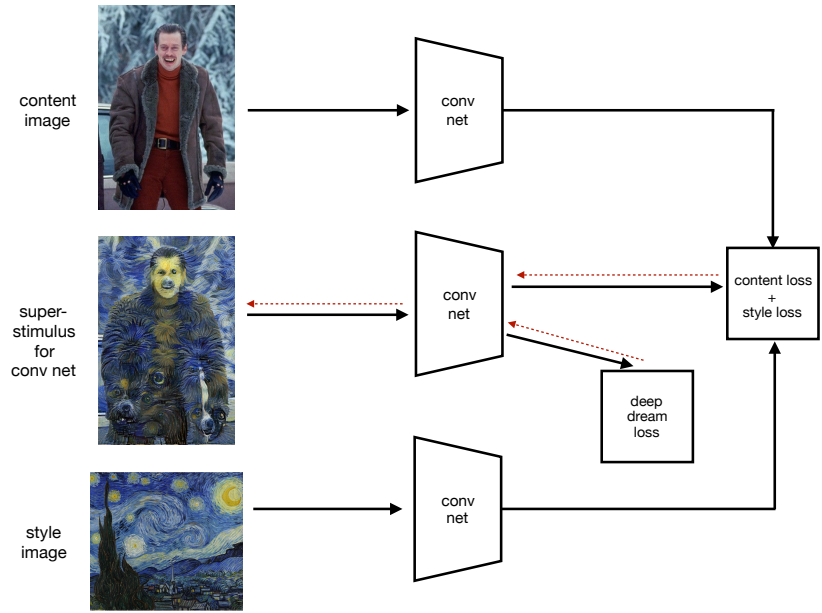}
\caption{Figure 12: Sensory Optimization for conv nets. Combining DD and
ST to generate a superstimulus for the conv net with different low-level
features than the content. Image generated by the author and titled ``A
Starry Night in Fargo''.}
\label{fig12}
\end{figure}

\begin{figure}
\centering
\includegraphics[width=0.99\textwidth,height=\textheight]{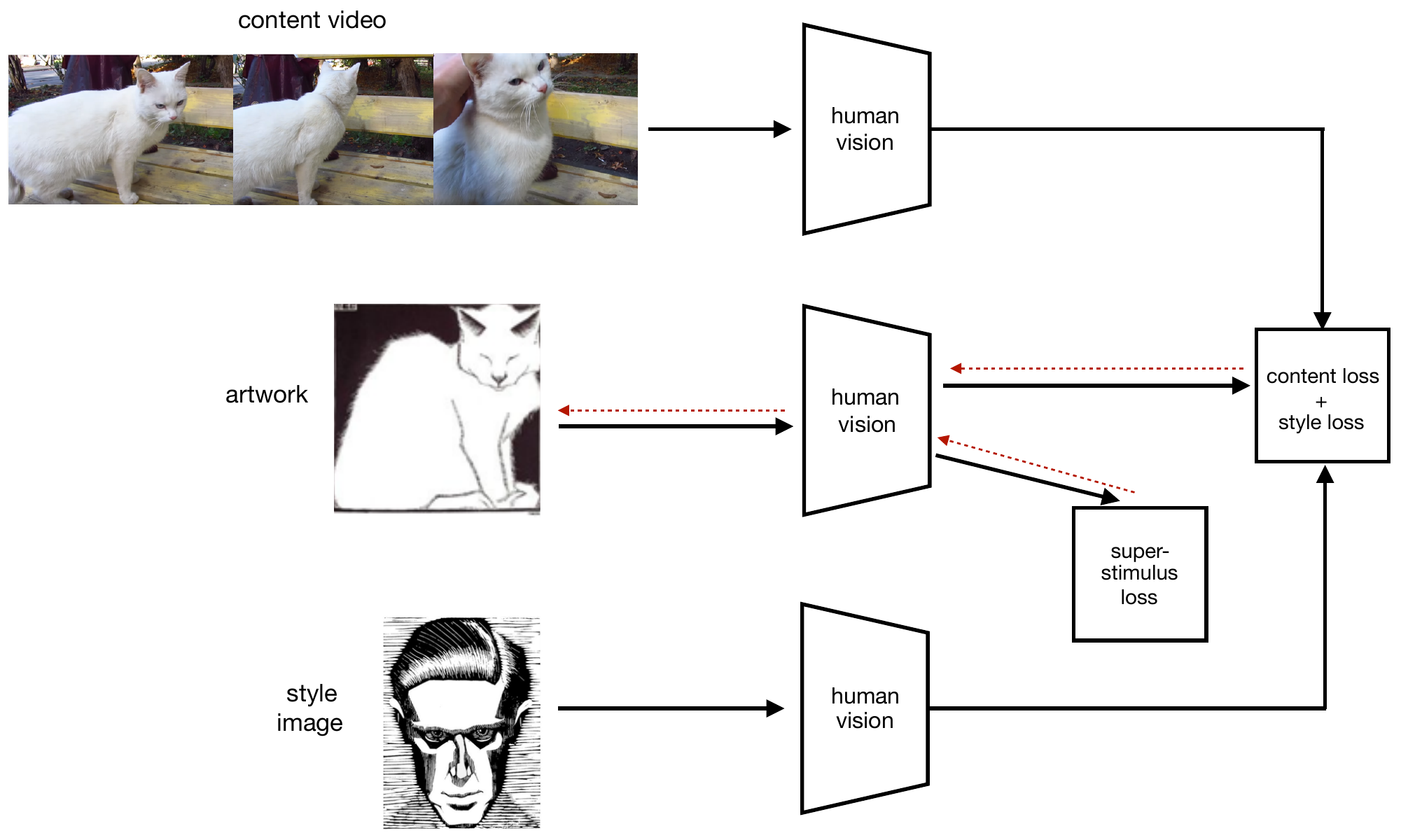}
\caption{Figure 13: Sensory Optimization for humans. The human visual
system (``human vision'') takes the role of conv net. The artwork is
optimized to be a superstimulus and to transcribe the content video into
the style of the style image. Constraints are also imposed by the
physical medium (as in Fig.~11) but this is not shown. For more on SO in
humans, see Section 2.2. \protect\hyperlink{sources}{Image sources.}}
\label{fig13}
\end{figure}

Combining FV and ST is one example of Sensory Optimization (see Fig.~\hyperref[fig12]{12}), and results of a simplified implementation of this are shown in
Figure \hyperref[fig10]{10}. However SO is not limited to artificial neural nets. Figure
\hyperref[fig13]{13} illustrates how humans could implement SO. This is similar to Style
Transfer (Fig.~\hyperref[fig8]{8}), but the content images (digital photos) are replaced
by a human's visual input (binocular video) and the loss functions
depend on the human's internal representation of this input. In Part 2,
SO in humans will be explained in detail.   

\begin{figure}
\centering
\includegraphics[width=0.9\textwidth,height=\textheight]{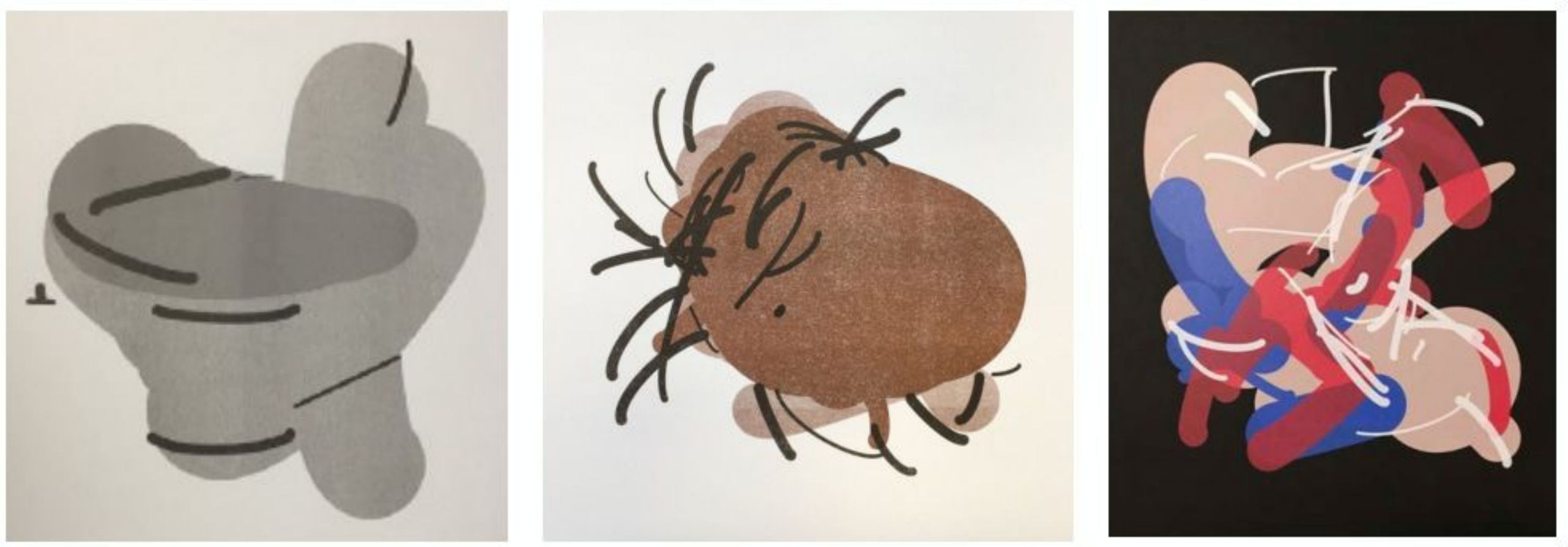}
\caption{Figure 14. Semi-abstract images that are classified as
``toilet'', ``house tick'', and ``pornographic'' (``NSFW'') by
recognition nets. From Tom White's ``Perception Engines'' and ``Pitch
Dream'' \citep{white2018perception}.}
\label{fig14}
\end{figure}

Another example of SO was introduced by the artist Tom White in his work
\href{https://medium.com/artists-and-machine-intelligence/perception-engines-8a46bc598d57}{``Perception
Engines''} \citep{white2018perception}. White used an algorithm to
discover abstract forms that look to a recognition net like everyday
objects. This combines FV and ST in one optimization objective. Even
more so than Deep Dream, the resulting images ``do a lot with a
little'', conveying an object class with a low complexity image (Fig.~\hyperref[fig14]{14}). White used the same idea to create abstract superstimuli for a
filter for pornographic images \citep{white2018synthetic}.\footnote{Strange
  quasi-pornographic \href{https://open_nsfw.gitlab.io/}{images} have
  also been generated using Feature Visualization \citep{goh}.} White's
abstract images often resemble (to human eyes) the objects they are
intended to depict. The fact that the neural net recognizes this same
resemblance is further evidence of generalization from photos to
stylized depictions. White also implemented a version of Medium Transfer
by making physical prints based on the digital images optimized by SO.
The recognition net classified the prints in the same way as the digital
images.

\hypertarget{part-2-sensory-optimization-and-human-cognition}{%
\subsection{Part 2: Sensory Optimization and human
cognition}\label{part-2-sensory-optimization-and-human-cognition}}

In the Introduction, I asked whether general human abilities are
sufficient for comprehending and creating visual art (``Generalist
position''). Part 1 showed that a neural net trained only on ImageNet
can be used to \emph{create} images that resemble human art. Given how
FV and ST work, if a net creates such images it must also
\emph{comprehend} the stylized representations in the images. Part 2
explores the implications of these neural net results for human
cognition and the Generalist position.

\hypertarget{the-sensory-optimization-hypothesis}{%
\subsubsection{2.1. The Sensory Optimization
Hypothesis}\label{the-sensory-optimization-hypothesis}}

I will formulate a hypothesis about how the results from Part 1 relate
to visual art in humans. The hypothesis has two premises:

\textbf{The Sensory Optimization Hypothesis}

\begin{enumerate}
\def\labelenumi{(\alph{enumi})}
\item
  If a system \(s\) has human-level visual recognition and can perform
  Sensory Optimization, then \(s\) can comprehend and create basic
  visual art.
\item
  Humans can perform Sensory Optimization.
\end{enumerate}

I'll clarify some terms. If system \(s\) ``performs Sensory
Optimization'' then \(s\) optimizes a physical artifact to cause
particular visual effects on \(s\). Neural nets can do this with
gradient descent, and in Section 2.2 I claim humans can do this using
general intelligence. The term ``basic visual art'' will be explained in
Section 2.3, but roughly means ``stylized, expressive representations in
a physical medium''.

The Sensory Optimization Hypothesis is a claim that the results from
neural nets in Part 1 generalize to humans. This claim has the following
implications:

\begin{enumerate}
\def\labelenumi{\arabic{enumi}.}
\item
  Visual art likely emerged once humans had the general intelligence to
  perform SO, i.e.~to create physical artifacts with the transcription
  and superstimulus properties. There was no need for a visual language
  or art-specific evolutionary adaptations.
\item
  The Style Transfer component of SO captures one way artists can borrow
  from previous work. This borrowing does not \emph{require} any
  art-specific abilities. (In practice, artists usually have extensive,
  art-specific knowledge and experience.)
\item
  We expect children to understand artistic representations from an
  early age. Adults in cultures without visual art should also be able
  to understand these representations with minimal instruction.
  Predictions for non-human primates are less clearcut.\footnote{As well
    as \citep{hochberg1962pictorial}, which was discussed in the
    Introduction, there is evidence that children can recognize objects
    in pictures before six months
    \citep{deloache1979picture, bloom2002children} suggesting this
    ability does not depend on verbal instruction. The literature on
    adults from cultures without visual art is more equivocal
    \citep{bovet2000picture}, but I'm unaware of any systematic evidence
    against the SO Hypothesis. There is also evidence showing that
    primates can spontaneously recognize objects in black-and-white
    photographs \citep{bovet2000picture}. Primates' ability to create
    visual art seems to be limited, but may be because they lack the
    general intelligence to perform SO.}
\end{enumerate}

In the following sections, I will address three objections to the
Sensory Optimization Hypothesis and in doing so clarify some of the
relevant concepts (``Sensory Optimization'', ``basic visual art'',
``human-level recognition''). Here are the objections:

\begin{quote}
\emph{2.2. Can humans perform Sensory Optimization?}\\
SO can be realized by the FV and ST algorithms applied to conv nets.
It's not clear the human brain can realize this kind of algorithm, as
our visual system differs from conv nets and we cannot optimize by
gradient descent. Moreover, the creative process in humans does not
resemble simple algorithms like FV.
\end{quote}

\begin{quote}
\emph{2.3. Can Sensory Optimization create impressive art?}\\
It's not clear that SO could ever create a set of artworks that matches
the range and depth of human art.
\end{quote}

\begin{quote}
\emph{2.4. Humans don't learn from photos}\\
The recognition nets in Part 1 are trained on photos. It's unclear
whether the results about FV and ST would carry over to nets trained on
data that is closer to human perceptual input (e.g.~something like
binocular video).
\end{quote}

\hypertarget{can-humans-perform-sensory-optimization}{%
\subsubsection{2.2. Can humans perform Sensory
Optimization?}\label{can-humans-perform-sensory-optimization}}

In neural nets, Sensory Optimization can be realized by combining
Feature Visualization and Style Transfer (Fig.~\hyperref[fig12]{12}). The human brain
cannot implement FV and ST but could implement \emph{analogs} of these
algorithms. This would involve evaluating analogs of the FV and ST
objective functions and then optimizing these objectives. I'll explain
the evaluation and optimization processes in turn.

\hypertarget{humans-evaluating-the-fv-and-st-objectives}{%
\paragraph{2.2.1. Humans evaluating the FV and ST
objectives}\label{humans-evaluating-the-fv-and-st-objectives}}

In FV the objective function for an image depends on the activation of
neurons in the recognition net. The neurons correspond to classes
(e.g.~dog), high-level features (e.g.~dog ears), and low-level features
(e.g.~shapes, colors). It's obvious that humans can recognize classes
and features in images. But are we aware (even subconsciously) of the
\emph{level} of activation of features in the brain? This is unclear.
However, we are aware of some quantities that seem to be related. People
can judge the \emph{typicality/prototypicality} of objects and features
and also judge how \emph{diagnostic} features are of a class.\footnote{People
  can look at a photo and judge whether someone is typical of a boxer or
  ballerina, and whether someone has typical ears or front teeth.
  Likewise, people are aware that a tiger's stripes are diagnostic of a
  tiger; the stripes distinguish the tiger from other animals. Also see
  \citep{murphy2004big, prototype}.} So people could optimize images
explicitly for typicality and diagnosticity. Moreover, when people
describe an image as vivid, beautiful or interesting, it's possible they
are partly responding to the typicality of elements of the
image.\footnote{It has been argued that beauty is associated with
  typicality and the vividness of caricature with diagnosticity
  \citep{reber2004processing, gombrich1960art}.}

In ST the objective is for the fusion image to match the representations
of the style and content images. Humans do not implement this exact
algorithm, but they do make similar evaluations. When shown two
paintings side-by-side, people without expertise in art can judge how
well they match in style.\footnote{Comparing paintings side-by-side is
  easier than identifying the style of a single painting.} This style
comparison is more demanding for humans than neural nets because we have
to switch visual attention between the two paintings.

\hypertarget{optimization-using-general-intelligence}{%
\paragraph{2.2.2. Optimization using general
intelligence}\label{optimization-using-general-intelligence}}

In neural nets, the optimization for Feature Visualization
simultaneously modifies millions of pixels every gradient step. Yet if
humans are modifying a painting or drawing, they take actions
\emph{serially}, one stroke at a time, which is a huge speed
disadvantage. Humans also evaluate images serially: many eye-movements
are required to take in a large, detailed painting
\citep{chater2018mind}.

How could humans do Sensory Optimization with this speed disadvantage?
It's important to note that SO in humans is blackbox search. When humans
first made art, they had no understanding of the human visual system and
no deterministic rules for drawing a picture of an animal that actually
looks like the animal. Humans were faced with a trial-and-error search,
where exploration is slow due to serial actions. Yet humans had a
crucial advantage: general intelligence. People can speed up search by
developing an intuition for which actions are promising (as in chess),
by hierarchical planning, by analytical techniques (e.g.~linear
perspective), and by learning from other people. This kind of
intelligent local search is plausibly how humans first created practical
technologies such as metal tools, architecture, and cooking.

The simplest application of SO in humans is to a single artist painting
a particular scene in a particular style. This matches the usual setup
for ST in neural nets. Yet this doesn't account either for the
development of style (see Section 2.2.2) or for how the content of
paintings depends on previous paintings. In painting a Madonna, medieval
artists would make only small changes in style and content to previous
Madonnas. This suggests a different ``implementation'' of SO in humans.
As well as being implemented by a single artist, SO can be implemented
by a succession of artists in an artistic lineage. Having many artists
contribute to the same enterprise allows for more exploration, which is
important given the slowness of human search.\footnote{Future research
  could explore versions of SO in neural nets with multiple distinct
  recognition nets. This is also relevant to explaining variation in
  tastes for art among humans. In neural nets, a single recognition net
  is used for SO but the ultimate consumers of the image are humans. If
  SO is implemented by multiple artists, there are multiple recognition
  nets, and the artists must have similar visual systems for this to
  work. At the same time, variation in visual systems among humans could
  account for some of the variation in tastes for visual art.}

\hypertarget{can-sensory-optimization-create-impressive-art}{%
\subsubsection{2.3. Can Sensory Optimization create impressive
art?}\label{can-sensory-optimization-create-impressive-art}}

Based on Part 1, Sensory Optimization appears to lack the following
features of human visual art:

\begin{enumerate}
\def\labelenumi{\arabic{enumi}.}
\item
  Coherence: e.g.~coherent faces, coherent 3D scenes, symmetrical forms.
\item
  Inventing new styles
\item
  Emotional expression and aesthetic properties (e.g.~beauty, elegance)
\item
  Communication of ideas, visual experiences, and narratives
\end{enumerate}

I believe SO can realize features 1-3, although not at the same level as
in human visual art. So the idea of ``basic visual art'' in the Sensory
Optimization Hypothesis is art that has features 1-3, as well as the
features of existing FV and ST images. I think SO will be less
successful at realizing feature 4.

\hypertarget{coherence}{%
\paragraph{2.3.1. Coherence}\label{coherence}}

In FV and DD images, individual objects are often structurally
incoherent (e.g.~asymmetric faces or bodies) and the overall arrangement
of objects is either stereotyped or random. One possible cause is the
local receptive fields of convolutional nets \citep{hinton2013taking}.
Another is that doing well on ImageNet does not require recognizing
whether objects or scenes are coherent. I expect general progress in
object recognition to mitigate these issues. So if SO is applied to
neural nets for vision in the future, I expect more coherent images.

\hypertarget{inventing-new-styles}{%
\paragraph{2.3.2. Inventing new styles}\label{inventing-new-styles}}

It's plausible that SO could be extended to develop artistic styles. The
idea would be to optimize the ``style image'' from ST as well as the
``fusion image''. As a concrete example, here's a possible objective
function:

\begin{itemize}
\item
  \emph{FV component}: We optimize for class labels related to outdoor
  scenes and landscapes (e.g.~trees, rivers, buildings, clouds,
  sunlight) and for positive emotions and beauty (see Section 2.3.3
  below).
\item
  \emph{ST component}: The content is outdoor scenes and the physical
  medium is oil paint. We also optimize over the style, by
  parameterizing the color palette, brush size, thickness of paint, etc.
\end{itemize}

I expect optimizing for this objective to yield a style that's closer to
Monet or Turner than to Egon Schiele, but it's hard to predict. The
general idea is that styles are optimized to help create superstimuli
for a certain kind of content. There are many ways to explore this idea
experimentally by building on existing work on FV and ST.

\hypertarget{emotional-expression-and-aesthetic-properties}{%
\paragraph{2.3.3. Emotional expression and aesthetic
properties}\label{emotional-expression-and-aesthetic-properties}}

The ability to communicate emotions and feelings is an important
property of art. For some theorists it's \emph{the} defining property of
high art \citep[
pp.264]{sep-collingwood-aesthetics, cooper1992companion}. Images
generated by FV and ST are lacking in both the range and clarity of
emotional expression. I claim this is because emotion labels were not
part of the recognition net's training. FV can optimize an image to
activate ``dog'' or ``dog snout'' neurons, but not to activate
``sadness'', ``fear'', ``peacefulness'', or ``nostalgia''.

In contrast to ImageNet, human visual experience is associated both with
object categories and with emotions. If you see a German Shepherd
running at you, you feel fear. If you see someone crying, you infer that
the person is sad. A dataset could be created where photos are annotated
with this kind of emotional association.\footnote{This would be related
  to existing datasets for image captioning, recognizing human actions,
  and recognizing emotions from facial expressions.} Training on such a
dataset would allow a neural net to predict both the emotions of people
in photos and also the emotions induced by scenes in the
photos.\footnote{It's not clear how well you can predict human emotional
  responses to abstract paintings from responses to photos of natural
  scenes. But this is a good topic to investigate.}

As well as emotions, aesthetic properties play a central role in visual
art. Works of art might be beautiful, sublime, elegant, uncanny, garish
or (intentionally) dull \citep[ pp.18]{davies2012artful}. It's not
possible to optimize for these properties having trained on ImageNet.
Yet human visual experience comes coupled with aesthetic evaluations. We
might see a beautiful tree, an elegant swan, an ugly patch of land. It's
plausible that training on a dataset with aesthetic annotations would
help SO generate images with particular aesthetic properties.\footnote{It's
  also plausible that fixing the problem of coherence from Section 2.3.1
  would help SO generate images with particular aesthetic properties.}

\hypertarget{communicating-ideas-and-experiences}{%
\paragraph{2.3.4. Communicating ideas and
experiences}\label{communicating-ideas-and-experiences}}

Consider a painting that depicts a scene from a particular war. Like a
news report, the painting straightforwardly communicates information
about the war. However, the painting could also be effective at putting
the viewer ``in the shoes'' of participants in the war, conveying the
feelings and thoughts of participants \citep{walton1990mimesis}. So the
viewer might learn something about this particular war and something
about war in general.

It does not seem possible for FV to generate this war painting from
scratch. FV generates images that reflect the \emph{existing} visual
knowledge of the recognition net. If the net doesn't know about the
particular war, then FV can't communicate either basic facts about the
war or rich experiences of it. (The war painting could be created by ST
with the appropriate content image. But this just passes the buck to
selecting the content image.)

Visual art can convey intellectual ideas, as is important in Surrealism,
Conceptual Art and many other movements. Visual art can also communicate
first-hand experiences of the world, as in the war painting example. It
seems difficult for SO to account for these features of art, and so this
would be a good topic for further research on SO.

\hypertarget{humans-dont-learn-from-photos}{%
\subsubsection{2.4. Humans don't learn from
photos}\label{humans-dont-learn-from-photos}}

The main evidence for the Sensory Optimization Hypothesis comes from
running Feature Visualization and Style Transfer on recognition nets
trained on ImageNet. The photos in ImageNet do not contain much visual
art. Nevertheless, the photos are arguably ``one step on the road
towards art''; they are static, rectangular, and were processed to make
them easy for humans to comprehend. This makes it less surprising that
FV/DD images resemble art and that ST works well.

Stronger evidence for the Sensory Optimization Hypothesis could be
obtained by training a network on inputs closer to human visual
perception. Historically, humans learned to recognize objects by seeing
them directly, and not by seeing photos or pictures. We could create a
dataset where the inputs are based on a raw binocular video stream, with
a shaky camera and varied points of view.

A related concern is that results in Part 1 from convolutional nets
would not generalize to other neural architectures (e.g.~recurrent
models, attention-based models). It would be valuable to see if FV and
ST can be ``ported'' to other architectures, especially those more
similar to the human visual system
\citep{sabour2017dynamic, ramachandran2019stand}.

\hypertarget{conclusion}{%
\subsection{Conclusion}\label{conclusion}}

Neural nets have great potential as a tool for understanding human
cognition for visual art. The training data and task objective for a net
can be precisely controlled, and we can learn about a trained net's
capabilities by probing its internal representations and testing
generalization on novel inputs.

In Part 1, I reviewed evidence about convolutional nets with minimal
exposure to visual art. Despite training to recognize objects, these
nets can be used to generate images that resemble visual art by the
process of Sensory Optimization. SO combines the ideas behind Feature
Visualization (superstimuli) and Style Transfer (transcription of
content). Images generated by SO can be construed as art ``for the
neural net'' rather than art for human consumption. I suggested ways to
extend SO in neural nets to capture more properties of human art
(Section 2.3). With richer datasets and tasks, SO could plausibly
generate images that have novel artistic styles and that are formally
coherent, emotionally expressive, and aesthetically pleasing.

I put forward the Sensory Optimization Hypothesis (Section 2.1). This is
claim that SO in neural nets illustrates a general phenomenon: if a
system can optimize artifacts to stimulate its own (human-level) visual
recognition system, then the system can make basic visual art. Humans
could optimize analogs of the FV and ST objectives by applying general
intelligence to speed up blackbox search. The Sensory Optimization
Hypothesis implies that general human abilities are sufficient for both
creating and developing human art, undermining arguments for the
necessity of cultural conventions (e.g.~a symbolic language) or an
innate art instinct.

There are many directions for future research:

\begin{itemize}
\item
  The Sensory Optimization Hypothesis could be tested more
  systematically, by looking at the images generated by networks with
  different architectures and more human-like training data (Section
  2.4).
\item
  SO could be investigated for visual art forms beyond painting and
  drawing, such as sculpture, architecture, fashion, cartoons, and
  movies.
\item
  Section 2.2.2 discussed how SO could help explain the historical
  development of visual art. To take this idea further, we could train a
  recognition net on both ImageNet and art up to a particular period and
  then investigate how this net interprets art from later periods.
  Building on Section 2.3.2, we could also investigate the kind of
  styles the net invents.
\item
  The idea underlying SO could be extended to music. Various people have
  taken a ``Generalist position'' on the cognitive abilities supporting
  music and have proposed theories on which music is a superstimulus for
  human language and auditory perception \citep[
  pp.138]{pinker2003how, davies2012artful}. It's also plausible that
  music ``transcribes'' various human internal experiences or emotional
  states into a different medium (namely a sequence of sounds). Given
  recent advances in neural nets for recognizing human speech, it might
  be possible to generate ``music for a neural net'', analogous to the
  ``art for neural nets'' generated by FV and ST.
\end{itemize}


\newpage

\hypertarget{acknowledgments}{%
\subsection*{Acknowledgments}\label{acknowledgments}}

Thanks to Peli Grietzer, Chris Olah, Justin Manley, Peter Insley, Anders
Sandberg, David Krueger, Brian Christian, and Asya Bergal for extremely
valuable discussions and feedback about the ideas in this work. I thank
Luba Elliott for inviting me to speak at the Creative AI Meetup, which
prompted some of the ideas in this paper. I also thank Peli Grietzer for
years of fruitful collaboration on related topics and for insightful
suggestions at various stages of this project.

\hypertarget{sources}{%
\subsection*{Image sources}\label{sources}}

\textbf{Figure 1 (top left to bottom right)}

\begin{itemize}
\item
  Sumerian warrior and bulls. Great Lyre of Ur. 2500BC.
\item
  Eastern Han Dynasty tomb fresco of chariots, horses, and men. Luoyang
  222-220AD
\item
  Ogata Korin. Duck and snow covered pine trees. 1700s
\item
  Master of Ikare. Carved Door (Ilekun), late 1800s. Brooklyn Museum.
\item
  L.S. Lowry. A Fight, Salford Art Gallery. 1935.
\item
  Pablo Picasso. Femme Assise Dora Maar. 1955.
\end{itemize}

\textbf{Figure 9 (wolf)}

\begin{itemize}
\item
  Linocut: Emil Nolde. Sail boat. 1906.
\item
  Oil painting: Gerhard Richter. Abstract Painting. 1997. High Museum of
  Art, Atlanta.
\item
  Oil painting: Vincent Van Gogh. Self portrait. 1889.
\item
  Abstract: Van der Leck. Composition No.~4. 1919.
\end{itemize}

\textbf{Figure 9 (dog)}

\begin{itemize}
\item
  Oil painting: Van Gogh. Olive Grove. 1889.
\item
  Abstract: Van der Leck. Composition No.~4. 1919.
\end{itemize}

\textbf{Figure 11:} MC Escher. Hand with fir cone. 1921

\textbf{Figure 13:} MC Escher. White Cat 1. 1919. MC Escher.
Self-portrait. 1919

\hypertarget{references}{%
\subsection*{}\label{references}}

  \bibliography{sensory_text.bib}

\begin{thebibliography}{10}

\bibitem{dutton2009art}
Denis Dutton.
\newblock {\em The art instinct: Beauty, pleasure, \& human evolution}.
\newblock Oxford University Press, USA, 2009.

\bibitem{goodman1976languages}
Nelson Goodman.
\newblock {\em Languages of art: An approach to a theory of symbols}.
\newblock Hackett publishing, 1976.

\bibitem{hyman2017depiction}
John Hyman and Katerina Bantinaki.
\newblock Depiction.
\newblock In Edward~N. Zalta, editor, {\em The Stanford Encyclopedia of
  Philosophy}. Metaphysics Research Lab, Stanford University, summer 2017
  edition, 2017.

\bibitem{pinker2003how}
Steven Pinker.
\newblock {\em How the mind works}.
\newblock Penguin UK, 2003.

\bibitem{ramachandran1999science}
Vilayanur~S Ramachandran and William Hirstein.
\newblock The science of art: A neurological theory of aesthetic experience.
\newblock {\em Journal of consciousness Studies}, 6(6-7):15--51, 1999.

\bibitem{gombrich1960art}
Ernst~H Gombrich.
\newblock Art and illusion: a study in the psychology of pictorial
  representation.
\newblock {\em New York: Pantheon}, 1960.

\bibitem{hochberg1962pictorial}
Julian Hochberg and Virginia Brooks.
\newblock Pictorial recognition as an unlearned ability: A study of one child's
  performance.
\newblock {\em The American Journal of Psychology}, 75(4):624--628, 1962.

\bibitem{zoph2018learning}
Barret Zoph, Vijay Vasudevan, Jonathon Shlens, and Quoc~V Le.
\newblock Learning transferable architectures for scalable image recognition.
\newblock In {\em Proceedings of the IEEE conference on computer vision and
  pattern recognition}, pages 8697--8710, 2018.

\bibitem{litjens2017survey}
Geert Litjens, Thijs Kooi, Babak~Ehteshami Bejnordi, Arnaud Arindra~Adiyoso
  Setio, Francesco Ciompi, Mohsen Ghafoorian, Jeroen~Awm Van Der~Laak, Bram
  Van~Ginneken, and Clara~I S{\'a}nchez.
\newblock A survey on deep learning in medical image analysis.
\newblock {\em Medical image analysis}, 42:60--88, 2017.

\bibitem{mordvintsev2015inceptionism}
Alexander Mordvintsev, Christopher Olah, and Mike Tyka.
\newblock Inceptionism: Going deeper into neural networks, 2015.
\newblock {\em Google Research Blog}, 2015.

\bibitem{gatys2016image}
Leon~A Gatys, Alexander~S Ecker, and Matthias Bethge.
\newblock Image style transfer using convolutional neural networks.
\newblock In {\em Proceedings of the IEEE conference on computer vision and
  pattern recognition}, pages 2414--2423, 2016.

\bibitem{deng2009imagenet}
Jia Deng, Wei Dong, Richard Socher, Li-Jia Li, Kai Li, and Li~Fei-Fei.
\newblock Imagenet: A large-scale hierarchical image database.
\newblock In {\em 2009 IEEE conference on computer vision and pattern
  recognition}, pages 248--255. Ieee, 2009.

\bibitem{chater2018mind}
Nick Chater.
\newblock {\em The mind is flat: The illusion of mental depth and the
  improvised mind}.
\newblock Penguin UK, 2018.

\bibitem{redmon2016you}
Joseph Redmon, Santosh Divvala, Ross Girshick, and Ali Farhadi.
\newblock You only look once: Unified, real-time object detection.
\newblock In {\em Proceedings of the IEEE conference on computer vision and
  pattern recognition}, pages 779--788, 2016.

\bibitem{redmon2018yolov3}
Joseph Redmon and Ali Farhadi.
\newblock Yolov3: An incremental improvement.
\newblock {\em arXiv preprint arXiv:1804.02767}, 2018.

\bibitem{olah2017feature}
Chris Olah, Alexander Mordvintsev, and Ludwig Schubert.
\newblock Feature visualization.
\newblock {\em Distill}, 2(11):e7, 2017.

\bibitem{nguyen2016multifaceted}
Anh Nguyen, Jason Yosinski, and Jeff Clune.
\newblock Multifaceted feature visualization: Uncovering the different types of
  features learned by each neuron in deep neural networks.
\newblock {\em arXiv preprint arXiv:1602.03616}, 2016.

\bibitem{santurkar2019computer}
Shibani Santurkar, Dimitris Tsipras, Brandon Tran, Andrew Ilyas, Logan
  Engstrom, and Aleksander Madry.
\newblock Computer vision with a single (robust) classifier.
\newblock {\em arXiv preprint arXiv:1906.09453}, 2019.

\bibitem{superstimulus}
{Wikipedia contributors}.
\newblock Supernormal stimulus --- {Wikipedia}{,} the free encyclopedia.
\newblock
  \url{https://en.wikipedia.org/w/index.php?title=Supernormal_stimulus&oldid=920961779"},
  2019.
\newblock [Online; accessed 22-October-2019].

\bibitem{dennett2006breaking}
Daniel~Clement Dennett.
\newblock {\em Breaking the spell: Religion as a natural phenomenon},
  volume~14.
\newblock Penguin, 2006.

\bibitem{davies2012artful}
Stephen Davies.
\newblock {\em The artful species: aesthetics, art, and evolution}.
\newblock OUP Oxford, 2012.

\bibitem{olah2018github}
{Olah, Chris}.
\newblock Research: Caricatures.
\newblock \url{https://github.com/tensorflow/lucid/issues/121}, 2018.
\newblock [Online; accessed 29-October-2019].

\bibitem{engstrom2019learning}
Logan Engstrom, Andrew Ilyas, Shibani Santurkar, Dimitris Tsipras, Brandon
  Tran, and Aleksander Madry.
\newblock Learning perceptually-aligned representations via adversarial
  robustness.
\newblock {\em arXiv preprint arXiv:1906.00945}, 2019.

\bibitem{mordvintsev2018differentiable}
Alexander Mordvintsev, Nicola Pezzotti, Ludwig Schubert, and Chris Olah.
\newblock Differentiable image parameterizations.
\newblock {\em Distill}, 2018.
\newblock https://distill.pub/2018/differentiable-parameterizations.

\bibitem{goodfellow2014explaining}
Ian~J Goodfellow, Jonathon Shlens, and Christian Szegedy.
\newblock Explaining and harnessing adversarial examples.
\newblock {\em arXiv preprint arXiv:1412.6572}, 2014.

\bibitem{brown2017adversarial}
Tom~B Brown, Dandelion Man{\'e}, Aurko Roy, Mart{\'\i}n Abadi, and Justin
  Gilmer.
\newblock Adversarial patch.
\newblock {\em arXiv preprint arXiv:1712.09665}, 2017.

\bibitem{white2018perception}
{White, Tom}.
\newblock Perception engines.
\newblock
  \url{https://medium.com/artists-and-machine-intelligence/perception-engines-8a46bc598d57},
  2018.
\newblock [Online; accessed 22-October-2019].

\bibitem{white2018synthetic}
{White, Tom}.
\newblock Synthetic abstractions.
\newblock
  \url{https://medium.com/@tom_25234/synthetic-abstractions-8f0e8f69f390},
  2018.
\newblock [Online; accessed 22-October-2019].

\bibitem{goh}
{Goh, Gabriel}.
\newblock Image synthesis from yahoo's open nsfw.
\newblock \url{https://open_nsfw.gitlab.io/}, 2016.
\newblock [Online; accessed 22-October-2019].

\bibitem{deloache1979picture}
Judy~S DeLoache, Mark~S Strauss, and Jane Maynard.
\newblock Picture perception in infancy.
\newblock {\em Infant Behavior and Development}, 2:77--89, 1979.

\bibitem{bloom2002children}
Paul Bloom.
\newblock {\em How children learn the meanings of words}.
\newblock MIT press, 2002.

\bibitem{bovet2000picture}
Dalila Bovet and Jacques Vauclair.
\newblock Picture recognition in animals and humans.
\newblock {\em Behavioural brain research}, 109(2):143--165, 2000.

\bibitem{murphy2004big}
Gregory Murphy.
\newblock {\em The big book of concepts}.
\newblock MIT press, 2004.

\bibitem{prototype}
{Wikipedia contributors}.
\newblock Prototype theory --- {Wikipedia}{,} the free encyclopedia.
\newblock
  \url{https://en.wikipedia.org/w/index.php?title=Prototype_theory&oldid=907628680"},
  2019.
\newblock [Online; accessed 22-October-2019].

\bibitem{reber2004processing}
R~Reber, N~Schwarz, and P~Winkielman.
\newblock Processing fluency and aesthetic pleasure: is beauty in the
  perceiver's processing experience?
\newblock {\em Personality and social psychology review: an official journal of
  the Society for Personality and Social Psychology, Inc}, 8(4):364--382, 2004.

\bibitem{hinton2013taking}
Geoffrey Hinton.
\newblock Taking inverse graphics seriously.
\newblock \url{http://www.csri.utoronto.ca/~hinton/csc2535/notes/lec6b.pdf},
  2013.

\bibitem{sep-collingwood-aesthetics}
Gary Kemp.
\newblock Collingwood's aesthetics.
\newblock In Edward~N. Zalta, editor, {\em The Stanford Encyclopedia of
  Philosophy}. Metaphysics Research Lab, Stanford University, fall 2016
  edition, 2016.

\bibitem{cooper1992companion}
David~Edward Cooper, Joseph Margolis, and Crispin Sartwell.
\newblock {\em A companion to aesthetics}.
\newblock Blackwell Oxford, UK, 1992.

\bibitem{walton1990mimesis}
Kendall~L Walton.
\newblock {\em Mimesis as make-believe: On the foundations of the
  representational arts}.
\newblock Harvard University Press, 1990.

\bibitem{sabour2017dynamic}
Sara Sabour, Nicholas Frosst, and Geoffrey~E Hinton.
\newblock Dynamic routing between capsules.
\newblock In {\em Advances in neural information processing systems}, pages
  3856--3866, 2017.

\bibitem{ramachandran2019stand}
Prajit Ramachandran, Niki Parmar, Ashish Vaswani, Irwan Bello, Anselm Levskaya,
  and Jonathon Shlens.
\newblock Stand-alone self-attention in vision models.
\newblock {\em arXiv preprint arXiv:1906.05909}, 2019.

\end{thebibliography}

\end{document}